\documentclass[11pt]{article}
\PassOptionsToPackage{table,xcdraw,dvipsnames}{xcolor}

\usepackage[final]{acl}

\usepackage{times}
\usepackage{latexsym}
\usepackage[T1]{fontenc}
\usepackage[utf8]{inputenc}
\usepackage{microtype}
\usepackage{inconsolata}

\usepackage{booktabs}
\usepackage{amsfonts}
\usepackage{amssymb}
\usepackage{nicefrac}
\usepackage{graphicx}
\usepackage{amsmath}
\usepackage{amsthm}
\usepackage{xcolor}
\usepackage{algorithm}
\usepackage{algorithmicx}
\usepackage{algpseudocode}
\usepackage{multirow}
\usepackage{subcaption}
\usepackage{makecell}
\usepackage{wrapfig}
\usepackage{enumitem}
\usepackage{todonotes}
\usepackage[most]{tcolorbox}
\usepackage{soul}
\soulregister{\texttt}{1}
\soulregister{\emph}{1}
\soulregister{\textit}{1}
\soulregister{\textbf}{1}

\definecolor{ecuMemory}{RGB}{184,217,232}    %
\definecolor{ecuStrategy}{RGB}{232,184,184}  %
\definecolor{ecuWorkflow}{RGB}{248,232,160}  %
\definecolor{ecuSkill}{RGB}{197,224,180}     %

\newcounter{prompt}[section]
\renewcommand{\theprompt}{\thesection.\arabic{prompt}}
\newcounter{casestudy}[section]
\renewcommand{\thecasestudy}{\thesection.\arabic{casestudy}}

\tcbset{
  uceframe/.style={
    enhanced,
    colback=gray!4,
    colframe=gray!45,
    boxrule=0.55pt,
    arc=3pt,
    boxsep=3pt,
    top=5pt,
    bottom=5pt,
    left=5pt,
    right=5pt,
    fonttitle=\bfseries\small,
    coltitle=black,
    colbacktitle=gray!12,
    attach boxed title to top left={yshift=-2mm, xshift=4mm},
    boxed title style={colframe=gray!45, colback=gray!12, sharp corners},
  },
  uceframefull/.style={uceframe, width=\linewidth},
  uceframewide/.style={uceframe, width=0.95\linewidth},
  ucememory/.style={uceframewide,
    colframe=ecuMemory!75!gray, colbacktitle=ecuMemory!60,
    boxed title style={colframe=ecuMemory!75!gray, colback=ecuMemory!60, sharp corners}},
  ucestrategy/.style={uceframewide,
    colframe=ecuStrategy!75!gray, colbacktitle=ecuStrategy!70,
    boxed title style={colframe=ecuStrategy!75!gray, colback=ecuStrategy!70, sharp corners}},
  uceworkflow/.style={uceframewide,
    colframe=ecuWorkflow!75!gray, colbacktitle=ecuWorkflow!60,
    boxed title style={colframe=ecuWorkflow!75!gray, colback=ecuWorkflow!60, sharp corners}},
  uceskill/.style={uceframewide,
    colframe=ecuSkill!75!gray, colbacktitle=ecuSkill!60,
    boxed title style={colframe=ecuSkill!75!gray, colback=ecuSkill!60, sharp corners}},
  uceinner/.style={
    enhanced,
    colback=white,
    colframe=gray!35,
    boxrule=0.4pt,
    arc=2pt,
    boxsep=2pt,
    left=4pt,
    right=4pt,
    top=3pt,
    bottom=3pt,
  },
  ucecomponent/.style={
    uceinner,
    colback=gray!6,
    top=2pt,
    bottom=2pt,
  },
}
\newcommand{\ucesuccess}[1]{{\sethlcolor{green!12}\hl{#1}}}
\newcommand{\ucefailure}[1]{{\sethlcolor{red!12}\hl{#1}}}

\theoremstyle{plain}

\theoremstyle{definition}

\theoremstyle{remark}

\title{Unified Context Evolution for LLM Agents}

\author{%
Zixuan Zhu$^{1,}$\thanks{Equal contribution}, Yitong Hu$^{2,*}$, Yong Dai$^{3}$, Junfeng Fang$^{4}$, Chunyang Jiang$^{5}$, \\\textbf{Senkang Hu}$^{6,}$\thanks{Corresponding author}, \textbf{Yuzhi Zhao}$^{7}$ \\
$^1$Nanyang Technological University,
$^2$Beijing University of Posts and Telecommunications, \\
$^3$X-Humanoid,
$^4$National University of Singapore, \\
$^5$Hong Kong University of Science and Technology,
$^6$City University of Hong Kong, \\
$^7$Huazhong University of Science and Technology \\
\texttt{zixuan012@e.ntu.edu.sg, senkang.forest@my.cityu.edu.hk}
}

\begin{document}
\maketitle

\begin{abstract}
LLM-based agents can solve multi-step interactive tasks by combining reasoning with environment feedback, yet each episode starts from the same fixed context and any useful strategy discovered along the way is lost once the task ends. Existing approaches either limit learning to the current task or pool all experience into a single untyped store, without distinguishing knowledge types, tracking quality through use, or balancing what the library still lacks. We introduce \emph{Unified Context Evolution} (UCE), a gradient-free framework that externalizes agent experience into an evolving library of typed \emph{Evolvable Context Units} (ECUs). UCE decomposes experience into four complementary types (Memory, Strategy, Workflow, and Skill), each generated from trajectories under type-specific conditions, retrieved at decision time, scored through repeated usage outcomes, and pruned when no longer valuable. A scheduling module allocates each cycle's generation budget toward the types where the library is weakest. Across two interactive benchmarks, UCE raises ALFWorld success from 75.4\% to 96.3\% and WebShop task score from 45.1\% to 61.3\%, and remains effective across alternative actor reasoning backbones without model retraining.
\end{abstract}

\section{Introduction}
\label{sec:intro}

Large language models have given rise to a new class of interactive agents that act over multiple steps to solve complex tasks. By interleaving reasoning with environment-grounded actions, an agent can carry out sequential decision tasks under environment feedback~\cite{yao2023react}, invoke external tools when its internal knowledge is insufficient~\cite{schick2023toolformer}, and explore alternative reasoning paths through structured search~\cite{yao2023tot}. Yet across these settings, each task is solved as an isolated event. The agent starts from the same fixed context (system instructions, few-shot demonstrations, tool schemas), and any mechanism it discovers during one episode disappears when the next task begins.

\begin{figure}[t]
  \centering
  \includegraphics[width=1\linewidth]{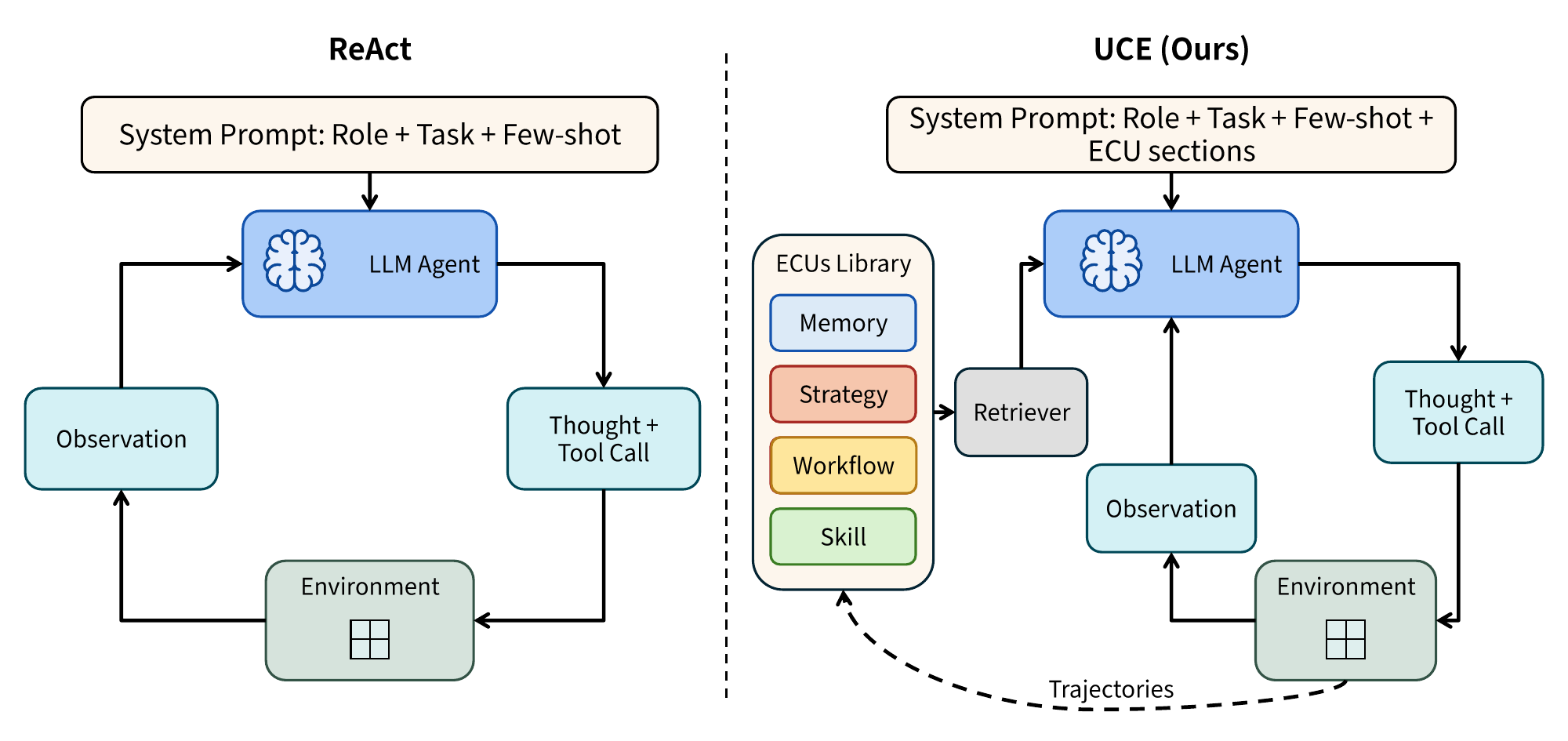}
  \caption{Comparison of the ReAct loop (left) and the UCE architecture (right).}
  \label{fig:comparison}
\end{figure}

Existing approaches address this limitation at different levels of persistence. Reinforcement learning methods train the agent's policy through environment feedback~\cite{xia2026skillrl,slearl2026}, improving capability but requiring gradient access and substantial cost. Within-task methods let the agent revise its approach during the current task through verbal self-critique~\cite{shinn2023reflexion} or iterative refinement~\cite{madaan2023selfrefine}, but the resulting improvements are confined to that task and lost when a new one begins. Beyond single-task adaptation, several systems keep the model frozen and accumulate prompt-injectable knowledge across tasks. Prior representations include mixed natural-language insights~\cite{zhao2024expel}, conditional guidelines~\cite{fu2024autoguide}, typed environmental manuals~\cite{chen2024automanual}, reusable workflows~\cite{wang2024awm}, and transferable or executable skills~\cite{nottingham2024sso,wang2023voyager,ren2026flowevo}; other work learns retrieval utility over stored episodes~\cite{zhang2026memrl}. These systems establish complementary pieces of persistent experience management. UCE addresses their joint organization: it separates four functional knowledge types with type-specific generation and retrieval boundaries, evaluates every unit through a common repeated-use fitness signal, and allocates generation resources across types as the library evolves.

\begin{table*}[t]
\centering
\footnotesize
\setlength{\tabcolsep}{5pt}
\renewcommand{\arraystretch}{1.12}
\begin{tabular}{@{}>{\raggedright\arraybackslash}p{0.12\textwidth}>{\raggedright\arraybackslash}p{0.23\textwidth}>{\raggedright\arraybackslash}p{0.20\textwidth}>{\raggedright\arraybackslash}p{0.34\textwidth}@{}}
\toprule
\textbf{Type} & \textbf{Defining question} & \textbf{Generation condition} & \textbf{Representative knowledge} \\
\midrule
\textbf{Memory} & How does an environment mechanism behave? & One trajectory & A sinkbasin can clean portable items without toggling a faucet. \\
\midrule
\textbf{Strategy} & What should the agent choose under a condition? & Multiple trajectories from the same task type & If \texttt{put} fails, open the receptacle and retry. \\
\midrule
\textbf{Workflow} & What execution sequence should the agent follow? & Successful trajectories from the same task type & Find the target, transform it, then place it at the target receptacle. \\
\midrule
\textbf{Skill} & How should a reusable operation be performed? & Trajectories from multiple task types & Search receptacles, open closed containers, and take the target when found. \\
\bottomrule
\end{tabular}
\caption{Four ECU types in UCE.}
\label{tab:ecu_types}
\end{table*}

We propose \textbf{Unified Context Evolution (UCE)} to make context itself the locus of learning. UCE maintains an evolving library of Evolvable Context Units (ECUs). Each ECU is a typed knowledge entry generated from collection trajectories, retrieved and injected into the actor prompt at decision time, evaluated through usage outcomes, and pruned when it becomes low-value or redundant. Figure~\ref{fig:comparison} contrasts the resulting architecture with a conventional ReAct loop. In ReAct each episode runs against a fixed prompt, whereas in UCE four retrieved ECU sections augment the actor prompt at every step and the library grows across cycles without updating the model weights. UCE can be instantiated with different actor reasoning backbones.

To make the gap concrete, consider two episodes from our experiments. In a household environment~\cite{shridhar2021alfworld}, an agent discovers after fourteen exploratory steps that a sinkbasin alone suffices to clean a portable item, with no faucet toggle or preliminary placement in the sink required. In an online shopping environment~\cite{yao2022webshop}, an agent clicks Buy Now from a read-only product sub-tab, receives repeated invalid-action errors, and exhausts its step budget without realizing that it must first navigate back to the main product page. Both mechanisms are reusable: the cleaning shortcut applies to every cleaning task, and the sub-tab navigation rule applies to every product page. Yet neither appears in the agent's few-shot demonstrations, and both discoveries vanish when the next task begins.

Our contributions are threefold:
\begin{itemize}[leftmargin=15pt,itemsep=2pt,topsep=2pt,parsep=0pt]
  \item We introduce \textbf{Evolvable Context Units (ECUs)}, a typed knowledge representation that decomposes agent experience into four functionally complementary types (Memory, Strategy, Workflow, Skill), each with independent generation conditions, retrieval boundaries, and elimination rules.
  \item We propose \textbf{Unified Context Evolution (UCE)}, a framework that manages the full ECU lifecycle through Knowledge Yield Scheduling and a usage-based fitness score.
  \item We evaluate UCE against several prompt-based agent methods on two interactive benchmarks, where the ECU library delivers stable gains across evolution cycles and UCE remains effective across four actor reasoning backbones.
\end{itemize}

\section{Related Work}
\label{sec:related}

\subsection{Reinforcement Learning for LLM Agents}

Reinforcement learning is a common route to improving LLM agents by updating model weights. WebRL~\cite{qi2024webrl} trains open-source web agents through a self-evolving online curriculum, and RAGEN~\cite{wang2025ragen} studies multi-turn agent RL in stochastic environments. Beyond direct policy training, HiPER~\cite{hiper2026} factors the policy into a high-level planner and a low-level executor for explicit credit assignment, SIOP~\cite{hu2026selfinducedoutcomepotentialturnlevel} studies verifier-free turn-level credit assignment, and retrieval-augmented methods use step-level experience retrieval or semantic information gain rewards~\cite{slearl2026,hu2026optimizingagenticreasoningretrieval}. SkillRL~\cite{xia2026skillrl} interleaves RL with the evolution of a reusable skill library, while SGSD~\cite{huang2026sgsd} uses an experience-derived skill-and-mistake bank as privileged context for outcome-gated self-distillation. MPO~\cite{xiong-etal-2025-mpo} instead trains a portable meta-planner with SFT and DPO to optimize high-level guidance from execution feedback. These approaches update either actor or auxiliary-model parameters and require training infrastructure. UCE trains neither the actor nor an auxiliary policy; it improves solely by evolving external context, while remaining compatible with closed-source actors.

\subsection{Within-Task Adaptation}

Closer to our setting, a parallel line of research keeps the model frozen and adapts the agent's behavior within the current task. ReAct~\cite{yao2023react} interleaves reasoning traces with environment actions and remains a common backbone for text-based agents. Reflexion~\cite{shinn2023reflexion} appends verbal self-reflections after failed trials and retries the same task. Self-Refine~\cite{madaan2023selfrefine} alternates generation and critique on a single instance. Other inference-time adaptation methods allocate a fixed search budget without model updates~\cite{fang2026inferencetimebudgetcontrolllm}. ReflAct~\cite{kim2025reflact} instead rewrites the ReAct thought format so that each step explicitly reflects on the current state relative to the task goal. These approaches alter how a frozen actor reasons or retries within an instance, but do not turn the resulting experience into persistent cross-task knowledge. UCE stores such learning as ECUs that later episodes can retrieve.

\subsection{Cross-Task Knowledge Libraries}

Most relevant to our work is a growing body of research that accumulates knowledge across tasks while keeping the actor frozen. ExpeL~\cite{zhao2024expel} extracts mixed natural-language insights into a flat pool, Self-Generated ICE~\cite{sarukkai2025ice} curates successful trajectories as in-context examples, and AutoGuide~\cite{fu2024autoguide} generates conditional guidelines and selects them for the current context. AutoManual~\cite{chen2024automanual} organizes environmental knowledge into typed rules that are updated online and compiled into a manual. Other systems emphasize particular knowledge functions: Mem0~\cite{chhikara2025mem0} maintains long-term conversational memory; ReasoningBank~\cite{ouyang2025reasoningbank} distills reasoning strategies from successful and failed experiences; Agent Workflow Memory~\cite{wang2024awm} induces reusable workflows; and ACE~\cite{zhang2025ace} evolves a curated context playbook. Voyager~\cite{wang2023voyager} and SkillWeaver~\cite{zheng2025skillweaver} build executable skill libraries, while Skill Set Optimization~\cite{nottingham2024sso} refines transferable in-context skills using downstream rewards. FlowEvo~\cite{ren2026flowevo} co-evolves workflows and executable skills and suppresses skills that cause negative transfer. MemRL~\cite{zhang2026memrl} learns retrieval utility for episodic memory through reinforcement learning. These works realize important components of persistent experience management, including typed rules, contextual retrieval, and usage-aware skill pruning. UCE combines four functionally complementary knowledge types in one library, gives each type distinct generation conditions and retrieval boundaries, evaluates every ECU with a common repeated-use fitness signal, and allocates generation budget across types via Knowledge Yield Scheduling.

\section{Methodology}
\label{sec:method}

\subsection{Problem Setup}

We consider an LLM agent operating in an interactive environment $\mathcal{E}$ on a task set $\mathcal{T} = \{t_1, \ldots, t_N\}$. Each task $t_i$ carries a type label $\tau_i \in \mathcal{Y}$. In each episode the agent receives observation $o_t$, produces action $a_t$, and the environment returns a new observation $o_{t+1}$. The episode ends when the task completes or the per-episode step limit is reached, producing either a binary success signal or a continuous task score depending on the environment. UCE partitions $\mathcal{T}$ into disjoint evaluation and collection subsets, ensuring that ECU content and fitness score are updated only from collection trajectories and never from evaluation trajectories.

\begin{figure*}[t]
  \centering
  \includegraphics[width=1\linewidth]{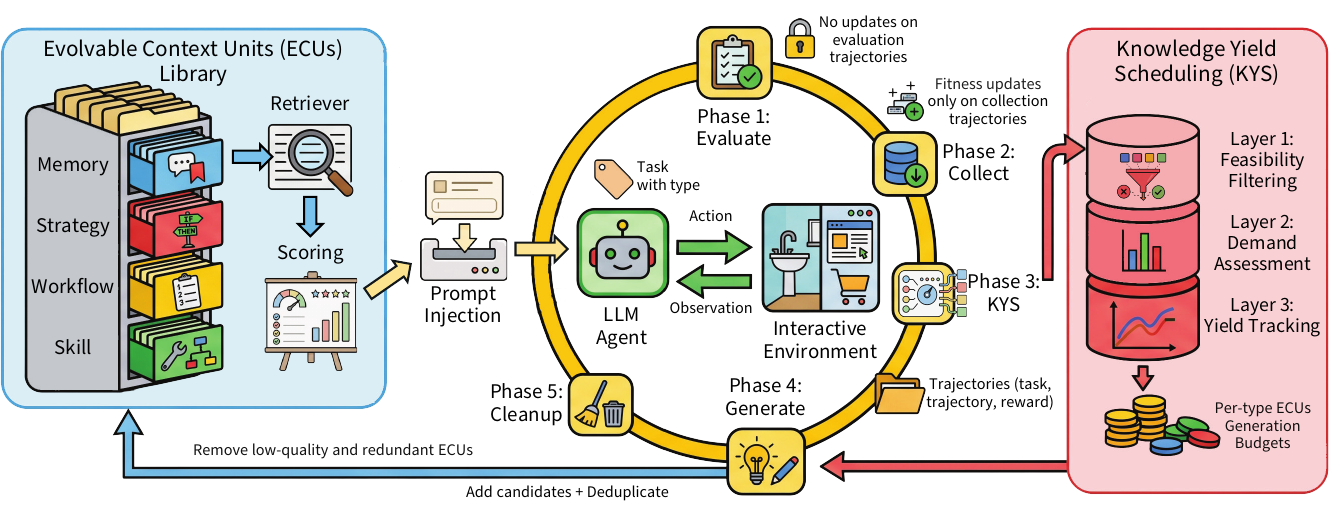}
  \caption{Overview of the UCE architecture. Five phases per cycle: evaluate, collect, KYS, generate, and cleanup.}
  \label{fig:architecture}
\end{figure*}

\subsection{ECU Library and Retrieval}

UCE maintains a structured knowledge library $\mathcal{L} = \{e_1, \ldots, e_M\}$ of ECUs. Each ECU is represented as a five-tuple:
\begin{equation}
    e = (z_e,\; \tau_e,\; w_e,\; c_e,\; f_e),
\end{equation}
Here $z_e \in \mathcal{Z}$ is the functional type, one of Memory, Strategy, Workflow, or Skill. The task type $\tau_e \in \mathcal{Y} \cup \{\emptyset\}$ records where the ECU applies, with Skill using $\emptyset$ to signal cross-type applicability. The field $w_e$ is a natural-language retrieval condition, used for matching but not injected into the prompt. The field $c_e$ is the knowledge content injected into the agent prompt, and $f_e$ is the fitness score defined in Section~\ref{sec:fitness}. Separating $w_e$ from $c_e$ lets retrieval optimize for relevance while injection exposes only actionable content to the agent.

\textbf{Four ECU types.}
The four ECU types are defined by the question they answer and the generation condition that produces them. Table~\ref{tab:ecu_types} summarizes the four types, with representative examples drawn from UCE runs.

Memory captures environment mechanisms discovered through trial-and-error within a single task instance, as a factual statement that asserts a property of the environment without prescribing any sequence of steps. Strategy encodes conditional decision rules at branching points within a task type, expressed in IF-THEN form. Workflow describes the canonical execution sequence for a task type, generalized so that it applies to all instances of that type. Skill specifies multi-step sub-operation methods that are reusable across task types. These four types coordinate at complementary granularities during agent execution. Memory explains the mechanisms behind specific environment interactions, Strategy guides the agent at decision points, Workflow provides the execution skeleton for a task type, and Skill expands specific steps within that skeleton.

\textbf{Retrieval.}
Before executing task $t_i$ (type $\tau_i$, natural-language description $d_i$), the retriever scores each ECU $e \in \mathcal{L}$:
\begin{equation}
\begin{aligned}
    &\mathrm{score}(e, t_i) = {}  \underbrace{\mathrm{sim}(\mathbf{h}_{w_e}, \mathbf{h}_{d_i})}_{\text{semantic similarity}} \\
    & + \underbrace{\beta \cdot \mathbf{1}[\tau_e = \tau_i]}_{\text{type-match boost}}  + \underbrace{\lambda \cdot \bigl(f_{\tau_i}(e) - 0.5\bigr)}_{\text{quality bias}},
\end{aligned}
\label{eq:score}
\end{equation}
Here $\mathbf{h}$ denotes sentence-transformer embeddings, and $\beta, \lambda > 0$ weight the task-type boost and quality bias. Before scoring, we apply a strict task-type filter. ECUs whose task type is nonempty and differs from the current task type are excluded, except for Skill ECUs, which are designed for cross-type reuse. Each ECU type is then independently ranked, and the top-$K$ per type are injected into four dedicated prompt sections, preventing any single type from crowding out the others.

\begin{table*}[t]
\centering
\setlength{\tabcolsep}{4pt}
\begin{tabular}{@{}l ccccccc cc@{}}
\toprule
\multirow{2}{*}{\textbf{Method}} & \multicolumn{7}{c}{\textbf{ALFWorld}} & \multicolumn{2}{c}{\textbf{WebShop}} \\
\cmidrule(lr){2-8} \cmidrule(lr){9-10}
 & Pick & Look & Clean & Heat & Cool & Pick2 & All & Score & Succ. \\
\midrule
ReAct                        & 83.3  & 50.0  & 96.8  & 78.3  & 71.4  & 52.9  & 75.4 & 45.1 & 30.0 \\
UCE, initial                 & 100.0 & 50.0  & 100.0 & 91.3  & 95.2  & 64.7  & 86.6 & 51.0 & 29.0 \\
\rowcolor{gray!15}
UCE, peak                    & 100.0 & 100.0 & 96.8  & 95.7  & 100.0 & 82.4  & 96.3 & 61.3 & 42.0 \\
\midrule
NoThinking                   & 91.7  & 27.8  & 87.1  & 73.9  & 81.0  & 58.8  & 73.1 & 58.7 & 37.0 \\
NoThinking + ECU, initial    & 95.8  & 38.9  & 93.5  & 78.3  & 76.2  & 70.6  & 78.4 & 65.6 & 40.0 \\
\rowcolor{gray!15}
NoThinking + ECU, peak       & 100.0 & 88.9  & 96.8  & 73.9  & 100.0 & 94.1  & 92.5 & 66.8 & 41.0 \\
\addlinespace
Plan-and-Act                 & 100.0 & 22.2  & 67.7  & 73.9  & 90.5  & 76.5  & 73.1 & 52.0 & 31.0 \\
Plan-and-Act + ECU, initial  & 95.8  & 33.3  & 100.0 & 82.6  & 81.0  & 76.5  & 81.3 & 56.8 & 37.0 \\
\rowcolor{gray!15}
Plan-and-Act + ECU, peak     & 95.8  & 94.4  & 100.0 & 100.0 & 100.0 & 94.1  & 97.8 & 62.7 & 42.0 \\
\addlinespace
ReflAct                      & 91.7  & 50.0  & 93.5  & 82.6  & 90.5  & 70.6  & 82.1 & 39.6 & 24.0 \\
ReflAct + ECU, initial       & 91.7  & 44.4  & 93.5  & 73.9  & 95.2  & 64.7  & 79.9 & 57.4 & 34.0 \\
\rowcolor{gray!15}
ReflAct + ECU, peak          & 95.8  & 100.0 & 96.8  & 87.0  & 100.0 & 76.5  & 93.3 & 61.5 & 39.0 \\
\midrule
ExpeL                        & 91.7  & 77.8  & 93.5  & 73.9  & 95.2  & 70.6  & 85.1 & 31.6 & 18.0 \\
ACE                          & 91.7  & 94.4  & 100.0 & 100.0 & 95.2  & 70.6  & 93.3 & 53.6 & 36.0 \\
AWM                          & 100.0 & 94.4  & 100.0 & 95.7  & 95.2  & 82.4  & 95.5 & 52.6 & 33.0 \\
ReasoningBank                & 95.8  & 61.1  & 87.1  & 100.0 & 90.5  & 70.6  & 85.8 & 44.9 & 26.0 \\
Reflexion                    & 66.7  & 94.4  & 51.6  & 52.2  & 95.2  & 35.3  & 64.9 & 69.0 & 45.0 \\
\bottomrule
\end{tabular}
\caption{Main results. ALFWorld reports per-task-type and overall success rate (\%). WebShop reports average task score (\%) and purchase success rate (\%).}
\label{tab:main}
\end{table*}
\begin{figure*}[t]
\centering
\includegraphics[width=\linewidth]{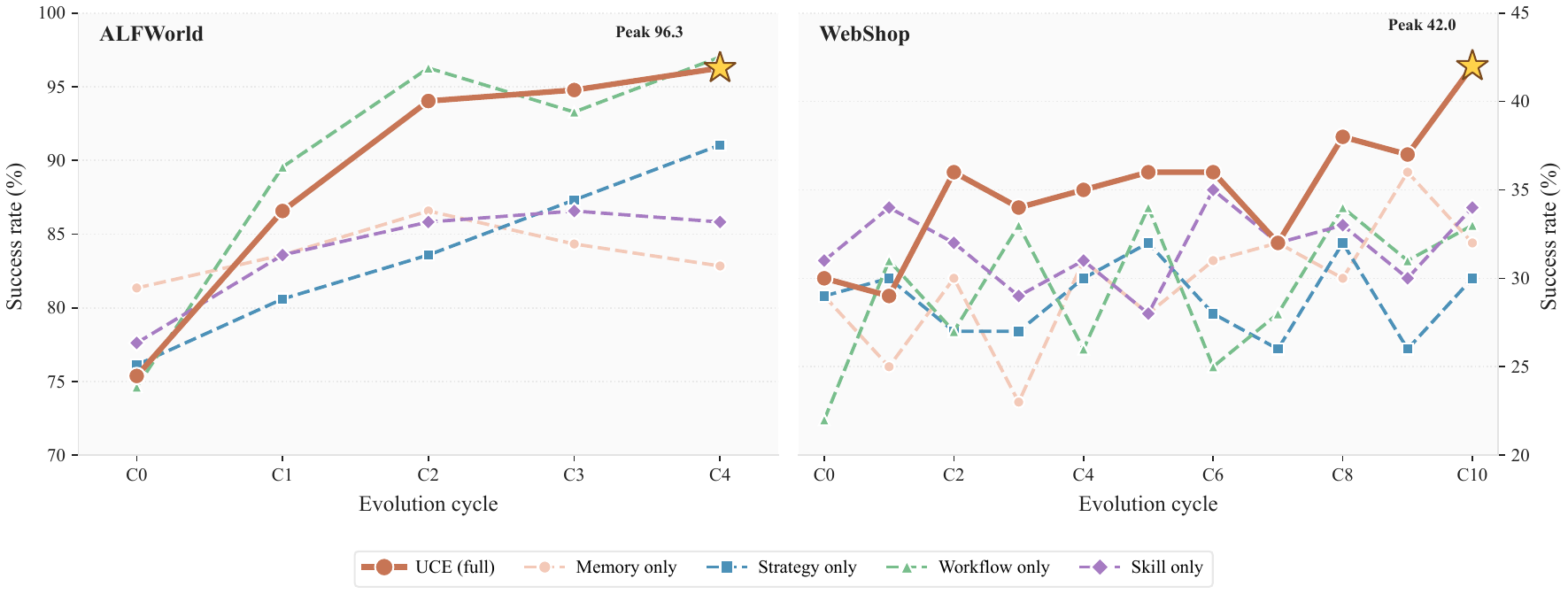}
\caption{Per-cycle success rate on ALFWorld (C0--C4) and WebShop (C0--C10). Solid lines show full UCE, dashed lines show single-type ablations, and the star marks the peak cycle. C0 is the ReAct baseline.}
\label{fig:evolution}
\end{figure*}

\subsection{Fitness Score and Lifecycle}
\label{sec:fitness}

Each ECU tracks its empirical contribution through a Laplace-smoothed fitness score:
\begin{equation}
    f_e = \frac{s_e + 1}{u_e + 2},
\end{equation}
Here $u_e$ is the cumulative usage count of $e$, and $s_e$ is the cumulative outcome credit. For binary-reward environments, $s_e$ increments by one on success and by zero on failure. For continuous-reward environments, $s_e$ increments by the final reward, so partial outcomes contribute fractional credit. A newly created ECU ($u_e = 0$) has $f_e = 0.5$, representing the prior belief that it is not yet verified.

For finer-grained quality assessment, we also maintain task-type-conditional statistics. When ECU $e$ has been used at least $u_{\mathrm{cond}}$ times on task type $\tau$, retrieval uses the conditional fitness score $f_\tau(e) = (s_\tau + 1)/(u_\tau + 2)$. Otherwise, retrieval falls back to the global fitness score $f_e$.

We define an \emph{adequate ECU} as one whose fitness score has been verified by sufficient usage and remains at or above the uninformative prior. Memory, Strategy, and Workflow generation skip task types already covered by an adequate ECU of the corresponding type.

\subsection{Evolution Cycle}
\label{sec:cycle}

Each evolution cycle $c$ executes five phases that transform the library from $\mathcal{L}_c$ to $\mathcal{L}_{c+1}$, as illustrated in Figure~\ref{fig:architecture}.

\textbf{Phase 1: Evaluate.}
The agent runs on the held-out evaluation task set with the current library $\mathcal{L}_c$ and records the success rate $r_c$. No fitness score is updated from these episodes, preserving the integrity of the evaluation signal.

\textbf{Phase 2: Collect.}
The agent runs on the disjoint collection task set with the same retrieval policy, storing each episode as a $(\text{task}, \text{trajectory}, \text{reward})$ tuple and recording the collection success rate $q_c$. Fitness scores are updated only from collection outcomes.

\textbf{Phase 3: Schedule.}
KYS reads the current library profile, the collection success rate $q_c$, the previous cycle's realized generation counts, and the nominal budget $B$, and produces integer per-type generation budgets $\{b_c^{(k)}\}_{k \in \mathcal{Z}}$ without mutating $\mathcal{L}_c$.

\textbf{Phase 4: Generate.}
For each ECU type with a nonzero budget, a type-specific generator prompt distills candidate ECUs from the collected trajectories. Each prompt includes quality feedback drawn from high- and low-fitness-score ECUs of the same type when available, together with a reshuffle test that rejects instance-specific observations (full prompts in Appendix~\ref{app:prompts}). Retained candidates are added to $\mathcal{L}_c$ after within-type deduplication.

\textbf{Phase 5: Cleanup.}
ECUs that have accumulated sufficient usage but whose fitness score falls below a threshold are removed. Within each (type, task\_type) group, only the highest-scoring ECU is kept once members have sufficient usage. The surviving ECUs form $\mathcal{L}_{c+1}$.

\subsection{Knowledge Yield Scheduling}
\label{sec:kys}

KYS exploits structural information already produced by the library. Generator skip conditions, task-type coverage, and fitness-score distributions are deterministic properties of $\mathcal{L}$ that can be read directly without additional computation. KYS organizes these signals into three layers, each providing a different view of the library's state.

\textbf{Layer 1: Feasibility filtering.}
The scheduler first determines whether each ECU type has remaining generation space. For Memory, Strategy, and Workflow, a type is feasible when at least one known task type lacks an adequate ECU of that type. Skill is always feasible. Infeasible types receive zero budget, which prevents saturated types from consuming generation calls.

\textbf{Layer 2: Demand assessment.}
For each feasible type, demand $d_c^{(k)}$ measures the current knowledge gap. Let $|\mathcal{Y}|$ denote the number of task types, and let $|\mathcal{Y}_{\mathrm{cov}}^{(k)}|$ denote the number of task types covered by adequate ECUs of type $k$. Memory and Strategy benefit from quality refinement even after initial coverage, so their demand combines a coverage gap with a quality gap:
\begin{equation}
    d_c^{(k)} =
    \Bigl(1 - \tfrac{|\mathcal{Y}_{\mathrm{cov}}^{(k)}|}{|\mathcal{Y}|}\Bigr)
    + \max\bigl(0,\, 0.5 - \bar{f}^{(k)}\bigr).
\end{equation}
Workflow needs one adequate procedure per task type, so demand measures coverage alone:
\begin{equation}
    d_c^{(k)} =
    \bigl(|\mathcal{Y}| - |\mathcal{Y}_{\mathrm{cov}}^{(k)}|\bigr) / |\mathcal{Y}|.
\end{equation}
Skill is cross-type, so demand is driven by the total count of adequate Skills relative to a soft reference $n_{\mathrm{ref}}$, with a floor to retain exploration:
\begin{equation}
    d_c^{(k)} =
    \max\bigl(0.1,\, 1 - n_{\mathrm{adeq}}^{(k)}/n_{\mathrm{ref}}\bigr).
\end{equation}

\textbf{Layer 3: Yield tracking.}
KYS maintains an exponential moving average of each type's historical return:
\begin{equation}
    y_c^{(k)} =
    (1 - \gamma)y_{c-1}^{(k)}
    + \gamma \Delta q_c
    \frac{g_{c-1}^{(k)}}{\sum_j g_{c-1}^{(j)}} ,
\end{equation}
where $\Delta q_c = q_c - q_{c-1}$ is the collection set success-rate change, and $g_{c-1}^{(k)}$ is the number of type $k$ ECUs generated in the previous cycle. The EMA is initialized to 0.5 for every ECU type. The yield update is skipped in the first cycle because no previous generation counts exist.

\textbf{Score fusion and budget assignment.}
For each feasible type, KYS computes
\begin{align}
    s_c^{(k)}
    &= \alpha_c d_c^{(k)} + (1 - \alpha_c)y_c^{(k)}, \\
    \alpha_c
    &= \max(\alpha_{\min}, \alpha_{\mathrm{start}} - \alpha_{\mathrm{decay}}c).
\end{align}
Scores are normalized over feasible types, then each feasible type receives a minimum share $\rho$. The remaining share is assigned in proportion to normalized scores and rounded to integer budgets summing to $B$. The realized number of added ECUs can still be lower than the assigned budget if the generator returns fewer valid candidates or deduplication removes near duplicates.

\section{Experiments}
\label{sec:experiment}

\subsection{Setup}

\begin{table*}[t]
\centering
\setlength{\tabcolsep}{4pt}
\begin{tabular}{@{}l l ccccccc cc@{}}
\toprule
\multirow{2}{*}{\textbf{ECU type}} & \multirow{2}{*}{\textbf{Snapshot}} & \multicolumn{7}{c}{\textbf{ALFWorld}} & \multicolumn{2}{c}{\textbf{WebShop}} \\
\cmidrule(lr){3-9} \cmidrule(lr){10-11}
 & & Pick & Look & Clean & Heat & Cool & Pick2 & All & Score & Succ. \\
\midrule
Memory   & Initial & 87.5  & 61.1 & 90.3  & 87.0  & 81.0  & 88.2  & 83.6 & 47.0 & 25.0 \\
Memory   & Peak    & 100.0 & 61.1 & 93.5  & 87.0  & 81.0  & 88.2  & 86.6 & 54.8 & 36.0 \\
\midrule
Strategy & Initial & 87.5  & 66.7 & 93.5  & 87.0  & 81.0  & 52.9  & 80.6 & 48.7 & 30.0 \\
Strategy & Peak    & 95.8  & 72.2 & 96.8  & 91.3  & 95.2  & 88.2  & 91.0 & 50.6 & 32.0 \\
\midrule
Workflow & Initial & 95.8  & 83.3 & 96.8  & 100.0 & 95.2  & 52.9  & 89.6 & 50.1 & 31.0 \\
Workflow & Peak    & 100.0 & 77.8 & 100.0 & 100.0 & 100.0 & 100.0 & 97.0 & 54.7 & 34.0 \\
\midrule
Skill    & Initial & 91.7  & 50.0 & 96.8  & 91.3  & 85.7  & 70.6  & 83.6 & 58.2 & 34.0 \\
Skill    & Peak    & 100.0 & 50.0 & 90.3  & 91.3  & 90.5  & 88.2  & 86.6 & 55.0 & 35.0 \\
\bottomrule
\end{tabular}
\caption{Single-type ablation. Each row retains only one ECU type in the library.}
\label{tab:ablation}
\end{table*}

\textbf{Benchmarks.}
We evaluate UCE on two text-based interactive benchmarks, ALFWorld~\cite{shridhar2021alfworld} and WebShop~\cite{yao2022webshop}. ALFWorld contains 134 household tasks across six task types. WebShop uses a 100-task split over Amazon product pages. Full benchmark statistics, task-type definitions, WebShop category mapping, and the $k{=}2$ stratified fold split are in Appendix~\ref{app:benchmarks}.

\textbf{Evaluation protocol.}
ALFWorld is run for five cycles and WebShop for eleven cycles. Cycle~0 is the ReAct baseline, and Cycle~$k$ reports performance after $k$ rounds of evolution. Each task is executed once per cycle with no multi-attempt retry.

\textbf{Models and hyperparameters.}
The acting agent is \texttt{gpt-4.1-mini}~\cite{openai2025gpt41} and the generator is \texttt{gpt-5.2}~\cite{openai2025gpt52}. Retrieval uses sentence-transformer embeddings~\cite{reimers2019sbert}. Full configuration is in Appendix~\ref{app:implementation}.

\textbf{Compared methods.}
We compare UCE against five cross-paradigm baselines: ExpeL~\cite{zhao2024expel}, Reflexion~\cite{shinn2023reflexion}, ACE~\cite{zhang2025ace}, AWM~\cite{wang2024awm}, and ReasoningBank~\cite{ouyang2025reasoningbank}. We also test UCE with three alternative actor backbones (NoThinking~\cite{ma2025nothinking,kim2025reflact}, Plan-and-Act~\cite{kim2025reflact}, and ReflAct~\cite{kim2025reflact}). All methods use the same actor model and, where applicable, generator/insight model, task sets, and step budgets. Per-method protocol details are deferred to Appendix~\ref{app:baseline_protocols}.

\subsection{Main Comparison}

Table~\ref{tab:main} reports the main results. On ALFWorld, UCE reaches 96.3\% success rate at its peak, a gain of 20.9 percentage points over the ReAct baseline. The largest improvements appear on the weakest baseline task types. Look rises from 50.0\% to 100.0\% and Pick2 from 52.9\% to 82.4\%. On WebShop, purchase success improves from 30.0\% to 42.0\% and task score from 45.1\% to 61.3\%.

Most ALFWorld gains arrive early, success jumps from 75.4\% at Cycle~0 to 86.6\% after the initial library and 94.0\% after one further update. The trajectory is not strictly monotonic for every task type, because each task is evaluated once per cycle under a newly retrieved library. Figure~\ref{fig:evolution} gives the full evolution curves; per-cycle tables for both benchmarks are in Appendix~\ref{app:trajectories}.

UCE remains effective across actor reasoning backbones. On ALFWorld, Plan-and-Act with UCE attains the overall best peak of 97.8\%, up from its 73.1\% no-library baseline. ReflAct with UCE improves from 82.1\% to 93.3\%, and NoThinking with UCE improves from 73.1\% to 92.5\%. On WebShop, task score improves by 8.1 to 21.9 percentage points across the three backbones. These results show that the UCE pipeline is not tied to the ReAct prompt format.

Among the cross-paradigm baselines, ExpeL reaches 85.1\% on ALFWorld but only 18.0\% purchase success on WebShop, where its flat insight pool provides less actionable guidance for the multi-step shopping flow. Among the cross-task knowledge baselines, AWM attains the highest ALFWorld success rate (95.5\%), while ACE performs best on WebShop (53.6 task score and 36.0\% success). Reflexion reaches 64.9\% on ALFWorld and 45.0\% on WebShop under a pass@3 protocol; its per-task multi-trial budget gives it a compute advantage on WebShop that single-trial methods do not share.

UCE's improvement over the ReAct baseline is statistically significant on both benchmarks: paired McNemar and bootstrap tests reject the null at $p < 0.05$, with all 95\% confidence intervals strictly above zero (Appendix~\ref{app:statsig}).

\begin{figure*}[t]
\centering
\begin{tcolorbox}[uceframefull, title=\textbf{Main-text case study: WebShop pillow-cover task}]
\small
\textbf{Task.} The agent must buy yellow, machine-washable Batmerry decorative pillow covers under \$50.
\par\smallskip\noindent
{\centering
\setlength{\tabcolsep}{4pt}
\begin{tabular}{@{}lcccp{0.42\linewidth}@{}}
\toprule
\textbf{Configuration} & \textbf{Steps} & \textbf{Reward} & \textbf{\#ECUs} & \textbf{Outcome} \\
\midrule
Full UCE & 9 & 1.00 & 3 & succeeds after recovering from a sub-tab \\
Memory only & 15 & 0.00 & 3 & invalid-action timeout \\
Strategy only & 15 & 0.00 & 3 & invalid-action timeout \\
Workflow only & 15 & 0.00 & 1 & invalid-action timeout \\
Skill only & 15 & 0.00 & 1 & invalid-action timeout \\
\bottomrule
\end{tabular}\par}
\par\smallskip\noindent
All configurations locate the relevant product page. The restricted runs click \texttt{Buy Now} while trapped inside the read-only \texttt{Features} sub-tab, receive repeated invalid-action feedback, and exhaust the step budget. Full UCE retrieves a workflow ECU for search and option selection, a memory ECU stating that product sub-tabs are read-only states, and a strategy ECU for returning with \texttt{< Prev}. After the first invalid \texttt{Buy Now}, the agent returns to the product page and completes the purchase. Case Study~\ref{case:pillow_cover} gives the full side-by-side trajectory.
\end{tcolorbox}
\caption{A WebShop task that requires multiple ECU types to succeed.}
\label{fig:main_case_webshop}
\end{figure*}

\subsection{Single-Type Ablation}

Table~\ref{tab:ablation} constrains the library to a single ECU type. On ALFWorld, Workflow alone reaches 97.0\%, close to the full library's 96.3\%, which indicates that reusable procedural skeletons explain the majority of the gain on this benchmark. Strategy peaks at 91.0\% and contributes at a finer decision-point granularity, while Memory (86.6\%) and Skill (86.6\%) provide smaller but complementary gains. On WebShop, the picture differs: no single type matches the full library, and the best single-type peak (Memory, 36.0\% success) remains 6.0 percentage points below the full library's 42.0\%. This gap reflects the multi-faceted nature of the shopping task, where the agent needs procedural workflow guidance, recovery strategies for navigation traps, and factual memory about page mechanisms to complete purchases reliably.

\subsection{Error Analysis and Case Study}

\begin{table}[t]
\centering
\setlength{\tabcolsep}{3pt}
\begin{tabular}{@{}lc@{}}
\toprule
\textbf{Failure mode} & \textbf{Share} \\
\midrule
Partial-mid attribute mismatch & 37\% \\
Max-step timeout & 27\% \\
Zero-match product failure & 15\% \\
Partial-high attribute mismatch & 13\% \\
Partial-low attribute mismatch & 8\% \\
\bottomrule
\end{tabular}
\caption{WebShop failure modes at Cycle~8.}
\label{tab:failure_modes_main}
\end{table}

Table~\ref{tab:failure_modes_main} categorizes the 62 failed WebShop episodes at Cycle~8. Half of them (partial-mid plus partial-high, 50\%) are attribute mismatches on a broadly correct product, where the agent identifies the right item type but selects a wrong size, color, or quantity option. These errors mark a natural boundary of prompt-level knowledge. UCE captures \emph{what to know} about environment mechanisms, decision rules, and procedures, while precise selection from a page-specific attribute panel primarily requires \emph{how to operate}, which is the strength of policy-optimizing reinforcement-learning methods. The next-largest category is max-step timeouts (27\%), which typically arise when the agent enters repeated search-click loops without committing to a purchase. Zero-match failures (15\%) reflect a different mode, where the agent commits to a product whose top-level category is wrong, indicating that the initial search query missed the intended item space. Partial-low failures (8\%) are residual cases where the agent identifies the right category but matches only a small fraction of the requested attributes. ALFWorld failures at the peak cycle (5 of 134) concentrate on Pick2 timeouts, the longest task type, where workflow and skill ECUs still leave room for improvement.

Figure~\ref{fig:main_case_webshop} illustrates a representative max-step timeout through a matched WebShop trajectory. The task is to purchase yellow, machine-washable Batmerry decorative pillow covers under \$50, and all five configurations issue the same opening search and successfully open the correct product page. From this shared state the trajectories diverge sharply. The four restricted libraries (Memory only, Strategy only, Workflow only, Skill only) each click into the \texttt{Features} sub-tab to inspect the listing, then attempt \texttt{Buy Now} while still inside the sub-tab. WebShop returns \texttt{Invalid action!}, and without any rule for recovering, the agent re-issues \texttt{Buy Now} repeatedly until the 15-step budget runs out with no purchase made. Full UCE retrieves three complementary ECUs that together resolve the same situation: a Workflow ECU encoding the search-then-purchase sequence, a Memory ECU recording the fact that product sub-tabs are read-only navigation states, and a Strategy ECU prescribing \texttt{< Prev} as the recovery action when an attempted purchase fails. After the first invalid \texttt{Buy Now}, the Strategy ECU triggers a \texttt{< Prev} that returns the agent to the main product page, where \texttt{Buy Now} succeeds with a reward of 1.0. This trajectory illustrates a general pattern visible across our Appendix case studies: when a single knowledge type drives the agent into a local trap, the surrounding ECU types supply the orthogonal information needed to escape it, and successful runs typically combine two or three types. Additional case studies are in Appendix~\ref{app:case_studies}.

\section{Conclusion}
\label{sec:conclusion}

We have presented Unified Context Evolution (UCE), a gradient-free framework in which an LLM agent improves across episodes by maintaining an evolving library of typed Evolvable Context Units. A usage-based fitness score tracks each unit's value, and Knowledge Yield Scheduling directs generation resources toward the weakest regions of the library. UCE raises ALFWorld success from 75.4\% to 96.3\% and WebShop task score from 45.1\% to 61.3\%, and remains effective across actor reasoning backbones.

Looking ahead, the four ECU types are currently hand-designed; an automatic type-discovery mechanism could adapt library structure to new domains. More broadly, coupling the fitness-driven lifecycle with lightweight parameter updates could yield agents that learn both what to retrieve and how to act on what they retrieve.

\section{Limitations}
\label{sec:limitations}

We evaluate UCE only on two text-based interactive environments with discrete action spaces. These results demonstrate its effectiveness on structured tasks, but do not yet establish UCE as a general-purpose self-evolving agent. Settings with multimodal observations, dynamic user or tool interactions, and substantially longer planning horizons, such as MobileWorld~\cite{kong2026mobileworld} and OSWorld~2.0~\cite{yuan2026osworld2}, remain untested, as do environments requiring continuous control. UCE also evolves knowledge within predefined structural boundaries: the four ECU types are manually specified, and task-type partitions are supplied or constructed in advance. This design makes generation and retrieval interpretable, but assumes that the relevant forms of experience and task organization are already known. UCE does not yet recognize when these representations are insufficient or create new structures beyond them. A more general system should be able to revise, extend, and reorganize these boundaries as its environment and experience change. Finally, several components of UCE still rely on hand-specified policies and thresholds. Future work should move toward agents that adapt not only the experience stored in context, but also the mechanisms used to organize, retrieve, and revise that experience.

\bibliography{ref}

\begin{thebibliography}{39}
\providecommand{\natexlab}[1]{#1}

\bibitem[{Chen et~al.(2024)Chen, Li, Yang, Yu, Lin, and
  He}]{chen2024automanual}
Minghao Chen, Yihang Li, Yanting Yang, Shiyu Yu, Binbin Lin, and Xiaofei He.
  2024.
\newblock \href {https://doi.org/10.52202/079017-0019} {{AutoManual}:
  Constructing instruction manuals by {LLM} agents via interactive
  environmental learning}.
\newblock In \emph{Advances in Neural Information Processing Systems},
  volume~37, pages 589--631. Curran Associates, Inc.

\bibitem[{Chhikara et~al.(2025)Chhikara, Khant, Aryan, Singh, and
  Yadav}]{chhikara2025mem0}
Prateek Chhikara, Dev Khant, Saket Aryan, Taranjeet Singh, and Deshraj Yadav.
  2025.
\newblock \href {https://doi.org/10.3233/FAIA251160} {Mem0: Building
  production-ready {AI} agents with scalable long-term memory}.
\newblock In \emph{ECAI 2025 -- 28th European Conference on Artificial
  Intelligence, 25--30 October 2025, Bologna, Italy -- Including 14th
  Conference on Prestigious Applications of Intelligent Systems (PAIS 2025)},
  volume 413 of \emph{Frontiers in Artificial Intelligence and Applications},
  pages 2993--3000. IOS Press.

\bibitem[{Fang et~al.(2026)Fang, Hu, Chang, Guo, Tao, Liu, Ruan, Huang, and
  Fang}]{fang2026inferencetimebudgetcontrolllm}
Zhengru Fang, Senkang~Forest Hu, Zhonghao Chang, Yu~Guo, Yihang Tao, Hongyao
  Liu, Mengzhe Ruan, Jun Huang, and Yuguang Fang. 2026.
\newblock \href {https://doi.org/10.48550/arXiv.2605.05701} {Inference-time
  budget control for {LLM} search agents}.
\newblock \emph{Preprint}, arXiv:2605.05701.

\bibitem[{Fu et~al.(2024)Fu, Kim, Kim, Sohn, Logeswaran, Bae, and
  Lee}]{fu2024autoguide}
Yao Fu, Dong-Ki Kim, Jaekyeom Kim, Sungryull Sohn, Lajanugen Logeswaran,
  Kyunghoon Bae, and Honglak Lee. 2024.
\newblock \href {https://doi.org/10.52202/079017-3811} {{AutoGuide}: Automated
  generation and selection of context-aware guidelines for large language model
  agents}.
\newblock In \emph{Advances in Neural Information Processing Systems},
  volume~37, pages 119919--119948. Curran Associates, Inc.

\bibitem[{Hu et~al.(2026{\natexlab{a}})Hu, Dai, Han, Fang, Zhao, Kwong, and
  Fang}]{hu2026selfinducedoutcomepotentialturnlevel}
Senkang Hu, Yong Dai, Xudong Han, Zhengru Fang, Yuzhi Zhao, Sam Tak~Wu Kwong,
  and Yuguang Fang. 2026{\natexlab{a}}.
\newblock \href {https://doi.org/10.48550/arXiv.2605.04984} {Self-induced
  outcome potential: Turn-level credit assignment for agents without
  verifiers}.
\newblock \emph{Preprint}, arXiv:2605.04984.

\bibitem[{Hu et~al.(2026{\natexlab{b}})Hu, Dai, Zhao, Tao, Guo, Fang, Kwong,
  and Fang}]{hu2026optimizingagenticreasoningretrieval}
Senkang Hu, Yong Dai, Yuzhi Zhao, Yihang Tao, Yu~Guo, Zhengru Fang, Sam Tak~Wu
  Kwong, and Yuguang Fang. 2026{\natexlab{b}}.
\newblock \href {https://openreview.net/forum?id=mUU7k4bN97} {Optimizing
  agentic reasoning with retrieval via synthetic semantic information gain
  reward}.
\newblock In \emph{Proceedings of the 43rd International Conference on Machine
  Learning (ICML)}.

\bibitem[{Huang et~al.(2026)Huang, Chen, Luo, Dai, Hu, and
  Zhao}]{huang2026sgsd}
Jiazhen Huang, Xiao Chen, Xiao Luo, Yong Dai, Senkang Hu, and Yuzhi Zhao. 2026.
\newblock \href {https://doi.org/10.48550/arXiv.2605.28791} {Skill-conditioned
  gated self-distillation for {LLM} reasoning}.
\newblock \emph{Preprint}, arXiv:2605.28791.

\bibitem[{Kagaya et~al.(2024)Kagaya, Yuan, Lou, Karlekar, Pranata, Kinose,
  Oguri, Wick, and You}]{kagaya2024rap}
Tomoyuki Kagaya, Thong~Jing Yuan, Yuxuan Lou, Jayashree Karlekar, Sugiri
  Pranata, Akira Kinose, Koki Oguri, Felix Wick, and Yang You. 2024.
\newblock \href {https://doi.org/10.48550/arXiv.2402.03610} {{RAP}:
  Retrieval-augmented planning with contextual memory for multimodal {LLM}
  agents}.
\newblock \emph{Preprint}, arXiv:2402.03610.

\bibitem[{Kim et~al.(2025)Kim, Rhee, Kim, Kim, Lee, Sung, and
  Jung}]{kim2025reflact}
Jeonghye Kim, Sojeong Rhee, Minbeom Kim, Dohyung Kim, Sangmook Lee, Youngchul
  Sung, and Kyomin Jung. 2025.
\newblock \href {https://doi.org/10.18653/v1/2025.emnlp-main.1697} {{ReflAct}:
  World-grounded decision making in {LLM} agents via goal-state reflection}.
\newblock In \emph{Proceedings of the 2025 Conference on Empirical Methods in
  Natural Language Processing}, pages 33433--33465, Suzhou, China. Association
  for Computational Linguistics.

\bibitem[{Kong et~al.(2026)Kong, Zhang, Yang, Gao, Liu, Tong, Cai, Zhou, Zhang,
  Chen, Liu, Hoi, and Wang}]{kong2026mobileworld}
Quyu Kong, Xu~Zhang, Zhenyu Yang, Nolan Gao, Chen Liu, Panrong Tong, Chenglin
  Cai, Hanzhang Zhou, Jianan Zhang, Liangyu Chen, Zhidan Liu, Steven Hoi, and
  Yue Wang. 2026.
\newblock \href {https://doi.org/10.18653/v1/2026.acl-long.278} {{MobileWorld}:
  Benchmarking autonomous mobile agents in agent-user interactive and
  {MCP}-augmented environments}.
\newblock In \emph{Proceedings of the 64th Annual Meeting of the Association
  for Computational Linguistics (Volume 1: Long Papers)}, pages 6142--6167, San
  Diego, California, United States. Association for Computational Linguistics.

\bibitem[{Ma et~al.(2025)Ma, He, Snell, Griggs, Min, and
  Zaharia}]{ma2025nothinking}
Wenjie Ma, Jingxuan He, Charlie Snell, Tyler Griggs, Sewon Min, and Matei
  Zaharia. 2025.
\newblock \href {https://doi.org/10.48550/arXiv.2504.09858} {Reasoning models
  can be effective without thinking}.
\newblock \emph{Preprint}, arXiv:2504.09858.

\bibitem[{Madaan et~al.(2023)Madaan, Tandon, Gupta, Hallinan, Gao, Wiegreffe,
  Alon, Dziri, Prabhumoye, Yang, Gupta, Majumder, Hermann, Welleck,
  Yazdanbakhsh, and Clark}]{madaan2023selfrefine}
Aman Madaan, Niket Tandon, Prakhar Gupta, Skyler Hallinan, Luyu Gao, Sarah
  Wiegreffe, Uri Alon, Nouha Dziri, Shrimai Prabhumoye, Yiming Yang, Shashank
  Gupta, Bodhisattwa~Prasad Majumder, Katherine Hermann, Sean Welleck, Amir
  Yazdanbakhsh, and Peter Clark. 2023.
\newblock \href {https://doi.org/10.52202/075280-2019} {Self-refine: Iterative
  refinement with self-feedback}.
\newblock In \emph{Advances in Neural Information Processing Systems},
  volume~36, pages 46534--46594. Curran Associates, Inc.

\bibitem[{Nottingham et~al.(2024)Nottingham, Majumder, Dalvi~Mishra, Singh,
  Clark, and Fox}]{nottingham2024sso}
Kolby Nottingham, Bodhisattwa~Prasad Majumder, Bhavana Dalvi~Mishra, Sameer
  Singh, Peter Clark, and Roy Fox. 2024.
\newblock \href {https://proceedings.mlr.press/v235/nottingham24a.html} {Skill
  set optimization: Reinforcing language model behavior via transferable
  skills}.
\newblock In \emph{Proceedings of the 41st International Conference on Machine
  Learning}, volume 235 of \emph{Proceedings of Machine Learning Research},
  pages 38409--38425. PMLR.

\bibitem[{{OpenAI}(2025{\natexlab{a}})}]{openai2025gpt41}
{OpenAI}. 2025{\natexlab{a}}.
\newblock \href {https://openai.com/index/gpt-4-1/} {Introducing {GPT-4.1} in
  the {API}}.

\bibitem[{{OpenAI}(2025{\natexlab{b}})}]{openai2025gpt52}
{OpenAI}. 2025{\natexlab{b}}.
\newblock \href {https://openai.com/index/introducing-gpt-5-2/} {Introducing
  {GPT-5.2}}.

\bibitem[{Ouyang et~al.(2026)Ouyang, Yan, Hsu, Chen, Jiang, Wang, Han, Le,
  Daruki, Tang, Tirumalashetty, Lee, Rofouei, Lin, Han, Lee, and
  Pfister}]{ouyang2025reasoningbank}
Siru Ouyang, Jun Yan, I-Hung Hsu, Yanfei Chen, Ke~Jiang, Zifeng Wang, Rujun
  Han, Long Le, Samira Daruki, Xiangru Tang, Vishy Tirumalashetty, George Lee,
  Mahsan Rofouei, Hangfei Lin, Jiawei Han, Chen-Yu Lee, and Tomas Pfister.
  2026.
\newblock \href {https://openreview.net/forum?id=jL7fwchScm} {{ReasoningBank}:
  Scaling agent self-evolving with reasoning memory}.
\newblock In \emph{The Fourteenth International Conference on Learning
  Representations}.

\bibitem[{Peng et~al.(2026)Peng, Liu, Zhou, Fleming, Wang, Garcia, and
  Hong}]{hiper2026}
Jiangweizhi Peng, Yuanxin Liu, Ruida Zhou, Charles Fleming, Zhaoran Wang,
  Alfredo Garcia, and Mingyi Hong. 2026.
\newblock \href {https://icml.cc/virtual/2026/poster/64058} {{HiPER}:
  Hierarchical plan--execute {RL} for multi-turn {LLM} agents}.
\newblock In \emph{Forty-Third International Conference on Machine Learning}.

\bibitem[{Qi et~al.(2025)Qi, Liu, Iong, Lai, Sun, Sun, Yang, Yang, Yao, Xu,
  Tang, and Dong}]{qi2024webrl}
Zehan Qi, Xiao Liu, Iat~Long Iong, Hanyu Lai, Xueqiao Sun, Jiadai Sun, Xinyue
  Yang, Yu~Yang, Shuntian Yao, Wei Xu, Jie Tang, and Yuxiao Dong. 2025.
\newblock \href
  {https://proceedings.iclr.cc/paper_files/paper/2025/hash/c66e1fcc9691aae706250638f36f681b-Abstract-Conference.html}
  {{WebRL}: Training {LLM} web agents via self-evolving online curriculum
  reinforcement learning}.
\newblock In \emph{International Conference on Learning Representations},
  volume 2025, pages 79791--79821.

\bibitem[{Reimers and Gurevych(2019)}]{reimers2019sbert}
Nils Reimers and Iryna Gurevych. 2019.
\newblock \href {https://doi.org/10.18653/v1/D19-1410} {Sentence-{BERT}:
  Sentence embeddings using {S}iamese {BERT}-networks}.
\newblock In \emph{Proceedings of the 2019 Conference on Empirical Methods in
  Natural Language Processing and the 9th International Joint Conference on
  Natural Language Processing (EMNLP-IJCNLP)}, pages 3982--3992, Hong Kong,
  China. Association for Computational Linguistics.

\bibitem[{Ren et~al.(2026)Ren, Yue, Li, Wang, Xu, Liu, Pan, and
  Di}]{ren2026flowevo}
Zeyu Ren, Ling Yue, Ran Li, Yishu Wang, Shengxiang Xu, Hanmo Liu, Shaowu Pan,
  and Shimin Di. 2026.
\newblock \href {https://openreview.net/forum?id=hU2N7IIkcE} {{FlowEvo}:
  Self-evolving agents through the co-evolution of workflows and executable
  skills}.
\newblock In \emph{Third Conference on Language Modeling}.

\bibitem[{Sarukkai et~al.(2025)Sarukkai, Xie, and Fatahalian}]{sarukkai2025ice}
Vishnu Sarukkai, Zhiqiang Xie, and Kayvon Fatahalian. 2025.
\newblock \href {https://doi.org/10.52202/085713-2158} {Self-generated
  in-context examples improve {LLM} agents for sequential decision-making
  tasks}.
\newblock In \emph{Advances in Neural Information Processing Systems},
  volume~38, pages 64392--64425. Curran Associates, Inc.

\bibitem[{Schick et~al.(2023)Schick, Dwivedi-Yu, Dessi, Raileanu, Lomeli,
  Hambro, Zettlemoyer, Cancedda, and Scialom}]{schick2023toolformer}
Timo Schick, Jane Dwivedi-Yu, Roberto Dessi, Roberta Raileanu, Maria Lomeli,
  Eric Hambro, Luke Zettlemoyer, Nicola Cancedda, and Thomas Scialom. 2023.
\newblock \href {https://doi.org/10.52202/075280-2997} {Toolformer: Language
  models can teach themselves to use tools}.
\newblock In \emph{Advances in Neural Information Processing Systems},
  volume~36, pages 68539--68551. Curran Associates, Inc.

\bibitem[{Shinn et~al.(2023)Shinn, Cassano, Gopinath, Narasimhan, and
  Yao}]{shinn2023reflexion}
Noah Shinn, Federico Cassano, Ashwin Gopinath, Karthik Narasimhan, and Shunyu
  Yao. 2023.
\newblock \href {https://doi.org/10.52202/075280-0377} {Reflexion: Language
  agents with verbal reinforcement learning}.
\newblock In \emph{Advances in Neural Information Processing Systems},
  volume~36, pages 8634--8652. Curran Associates, Inc.

\bibitem[{Shridhar et~al.(2021)Shridhar, Yuan, C{\^o}t{\'e}, Bisk, Trischler,
  and Hausknecht}]{shridhar2021alfworld}
Mohit Shridhar, Xingdi Yuan, Marc-Alexandre C{\^o}t{\'e}, Yonatan Bisk, Adam
  Trischler, and Matthew Hausknecht. 2021.
\newblock \href {https://openreview.net/forum?id=0IOX0YcCdTn} {{ALFWorld}:
  Aligning text and embodied environments for interactive learning}.
\newblock In \emph{International Conference on Learning Representations}.

\bibitem[{Wang et~al.(2024)Wang, Xie, Jiang, Mandlekar, Xiao, Zhu, Fan, and
  Anandkumar}]{wang2023voyager}
Guanzhi Wang, Yuqi Xie, Yunfan Jiang, Ajay Mandlekar, Chaowei Xiao, Yuke Zhu,
  Linxi Fan, and Anima Anandkumar. 2024.
\newblock \href {https://openreview.net/forum?id=ehfRiF0R3a} {Voyager: An
  open-ended embodied agent with large language models}.
\newblock \emph{Transactions on Machine Learning Research}.

\bibitem[{Wang et~al.(2023)Wang, Xu, Lan, Hu, Lan, Lee, and
  Lim}]{wang2023plansolve}
Lei Wang, Wanyu Xu, Yihuai Lan, Zhiqiang Hu, Yunshi Lan, Roy Ka-Wei Lee, and
  Ee-Peng Lim. 2023.
\newblock \href {https://doi.org/10.18653/v1/2023.acl-long.147} {Plan-and-solve
  prompting: Improving zero-shot chain-of-thought reasoning by large language
  models}.
\newblock In \emph{Proceedings of the 61st Annual Meeting of the Association
  for Computational Linguistics (Volume 1: Long Papers)}, pages 2609--2634,
  Toronto, Canada. Association for Computational Linguistics.

\bibitem[{Wang and Jiang(2026)}]{slearl2026}
Prince~Zizhuang Wang and Shuli Jiang. 2026.
\newblock \href {https://doi.org/10.48550/arXiv.2603.18079} {{SLEA-RL}:
  Step-level experience augmented reinforcement learning for multi-turn agentic
  training}.
\newblock \emph{Preprint}, arXiv:2603.18079.

\bibitem[{Wang et~al.(2025{\natexlab{a}})Wang, Wang, Wang, Zhang, Li, Yang,
  Jin, Yu, Nguyen, Liu, Gottlieb, Lu, Cho, Wu, Li, Wang, Choi, and
  Li}]{wang2025ragen}
Zihan Wang, Kangrui Wang, Qineng Wang, Pingyue Zhang, Linjie Li, Zhengyuan
  Yang, Xing Jin, Kefan Yu, Minh~Nhat Nguyen, Licheng Liu, Eli Gottlieb, Yiping
  Lu, Kyunghyun Cho, Jiajun Wu, Fei-Fei Li, Lijuan Wang, Yejin Choi, and
  Manling Li. 2025{\natexlab{a}}.
\newblock \href {https://doi.org/10.48550/arXiv.2504.20073} {{RAGEN}:
  Understanding self-evolution in {LLM} agents via multi-turn reinforcement
  learning}.
\newblock \emph{Preprint}, arXiv:2504.20073.

\bibitem[{Wang et~al.(2025{\natexlab{b}})Wang, Mao, Fried, and
  Neubig}]{wang2024awm}
Zora~Zhiruo Wang, Jiayuan Mao, Daniel Fried, and Graham Neubig.
  2025{\natexlab{b}}.
\newblock \href {https://proceedings.mlr.press/v267/wang25bx.html} {Agent
  workflow memory}.
\newblock In \emph{Proceedings of the 42nd International Conference on Machine
  Learning}, volume 267 of \emph{Proceedings of Machine Learning Research},
  pages 63897--63911. PMLR.

\bibitem[{Xia et~al.(2026)Xia, Chen, Wang, Liu, Zeng, Wang, Han, Zhou, Zhao,
  Chen, Zheng, Xie, and Yao}]{xia2026skillrl}
Peng Xia, Jianwen Chen, Hanyang Wang, Jiaqi Liu, Kaide Zeng, Yu~Wang, Siwei
  Han, Yiyang Zhou, Xujiang Zhao, Haifeng Chen, Zeyu Zheng, Cihang Xie, and
  Huaxiu Yao. 2026.
\newblock \href {https://doi.org/10.48550/arXiv.2602.08234} {{SkillRL}:
  Evolving agents via recursive skill-augmented reinforcement learning}.
\newblock \emph{Preprint}, arXiv:2602.08234.

\bibitem[{Xiong et~al.(2025)Xiong, Song, Dong, Zhao, Song, XWang, and
  Li}]{xiong-etal-2025-mpo}
Weimin Xiong, Yifan Song, Qingxiu Dong, Bingchan Zhao, Feifan Song, XWang, and
  Sujian Li. 2025.
\newblock \href {https://doi.org/10.18653/v1/2025.findings-emnlp.210} {{MPO}:
  Boosting {LLM} agents with meta plan optimization}.
\newblock In \emph{Findings of the Association for Computational Linguistics:
  EMNLP 2025}, pages 3914--3935, Suzhou, China. Association for Computational
  Linguistics.

\bibitem[{Yao et~al.(2022)Yao, Chen, Yang, and Narasimhan}]{yao2022webshop}
Shunyu Yao, Howard Chen, John Yang, and Karthik Narasimhan. 2022.
\newblock \href {https://doi.org/10.52202/068431-1508} {{WebShop}: Towards
  scalable real-world web interaction with grounded language agents}.
\newblock In \emph{Advances in Neural Information Processing Systems},
  volume~35, pages 20744--20757. Curran Associates, Inc.

\bibitem[{Yao et~al.(2023{\natexlab{a}})Yao, Yu, Zhao, Shafran, Griffiths, Cao,
  and Narasimhan}]{yao2023tot}
Shunyu Yao, Dian Yu, Jeffrey Zhao, Izhak Shafran, Tom Griffiths, Yuan Cao, and
  Karthik Narasimhan. 2023{\natexlab{a}}.
\newblock \href {https://doi.org/10.52202/075280-0517} {Tree of thoughts:
  Deliberate problem solving with large language models}.
\newblock In \emph{Advances in Neural Information Processing Systems},
  volume~36, pages 11809--11822. Curran Associates, Inc.

\bibitem[{Yao et~al.(2023{\natexlab{b}})Yao, Zhao, Yu, Du, Shafran, Narasimhan,
  and Cao}]{yao2023react}
Shunyu Yao, Jeffrey Zhao, Dian Yu, Nan Du, Izhak Shafran, Karthik~R.
  Narasimhan, and Yuan Cao. 2023{\natexlab{b}}.
\newblock \href {https://openreview.net/forum?id=WE_vluYUL-X} {{ReAct}:
  Synergizing reasoning and acting in language models}.
\newblock In \emph{The Eleventh International Conference on Learning
  Representations}.

\bibitem[{Yuan et~al.(2026)Yuan, Zhou, Xiong, Wu, Sun, Song, Cui, Wang, Wu, Li,
  Lu, Lu, Zhen, Wang, Deng, Yang, Chen, Zheng, Su, Yu, Zou, Agashe, Lu, Kaur,
  Qi, Chen, Sala, Liu, Lin, Yu, Su, Reddy, Wang, Qi, Xie, and
  Yu}]{yuan2026osworld2}
Mengqi Yuan, Zilong Zhou, Xinzhuang Xiong, Weiming Wu, Jiayang Sun, Jiamin
  Song, Kaiqian Cui, Bowen Wang, Haoyuan Wu, Yitong Li, Dunjie Lu, Haikong Lu,
  Qi~Zhen, Xinyuan Wang, Jiaqi Deng, Yuhao Yang, Cheng Chen, Boyuan Zheng, Alex
  Su, and 17 others. 2026.
\newblock \href {https://doi.org/10.48550/arXiv.2606.29537} {{OSWorld} 2.0:
  Benchmarking computer use agents on long-horizon real-world tasks}.
\newblock \emph{Preprint}, arXiv:2606.29537.

\bibitem[{Zhang et~al.(2026{\natexlab{a}})Zhang, Hu, Upasani, Ma, Hong,
  Kamanuru, Rainton, Wu, Ji, Li, Thakker, Zou, and Olukotun}]{zhang2025ace}
Qizheng Zhang, Changran Hu, Shubhangi Upasani, Boyuan Ma, Fenglu Hong,
  Vamsidhar Kamanuru, Jay Rainton, Chen Wu, Mengmeng Ji, Hanchen Li, Urmish
  Thakker, James~Y Zou, and Kunle Olukotun. 2026{\natexlab{a}}.
\newblock \href
  {https://proceedings.iclr.cc/paper_files/paper/2026/hash/8a94ff6f922d995d7d3f4ebf4143e442-Abstract-Conference.html}
  {Agentic context engineering: Evolving contexts for self-improving language
  models}.
\newblock In \emph{The Fourteenth International Conference on Learning
  Representations}.

\bibitem[{Zhang et~al.(2026{\natexlab{b}})Zhang, Wang, Zhou, Liao, Feng, Li,
  Zheng, Zhang, Wen, Li, Xiong, Qi, Tang, and Wen}]{zhang2026memrl}
Shengtao Zhang, Jiaqian Wang, Ruiwen Zhou, Junwei Liao, Yuchen Feng, Zhuo Li,
  Yujie Zheng, Weinan Zhang, Ying Wen, Zhiyu Li, Feiyu Xiong, Yutao Qi,
  Bo~Tang, and Muning Wen. 2026{\natexlab{b}}.
\newblock \href {https://doi.org/10.48550/arXiv.2601.03192} {{MemRL}:
  Self-evolving agents via runtime reinforcement learning on episodic memory}.
\newblock \emph{Preprint}, arXiv:2601.03192.

\bibitem[{Zhao et~al.(2024)Zhao, Huang, Xu, Lin, Liu, and
  Huang}]{zhao2024expel}
Andrew Zhao, Daniel Huang, Quentin Xu, Matthieu Lin, Yong-Jin Liu, and Gao
  Huang. 2024.
\newblock \href {https://doi.org/10.1609/aaai.v38i17.29936} {{ExpeL}: {LLM}
  agents are experiential learners}.
\newblock \emph{Proceedings of the AAAI Conference on Artificial Intelligence},
  38(17):19632--19642.

\bibitem[{Zheng et~al.(2025)Zheng, Fatemi, Jin, Wang, Gandhi, Song, Gu,
  Srinivasa, Liu, Neubig, and Su}]{zheng2025skillweaver}
Boyuan Zheng, Michael~Y. Fatemi, Xiaolong Jin, Zora~Zhiruo Wang, Apurva Gandhi,
  Yueqi Song, Yu~Gu, Jayanth Srinivasa, Gaowen Liu, Graham Neubig, and Yu~Su.
  2025.
\newblock \href {https://doi.org/10.48550/arXiv.2504.07079} {{SkillWeaver}: Web
  agents can self-improve by discovering and honing skills}.
\newblock \emph{Preprint}, arXiv:2504.07079.

\end{thebibliography}

\appendix
\newpage
\section{Benchmark Details}
\label{app:benchmarks}

\subsection{ALFWorld}

ALFWorld~\cite{shridhar2021alfworld} is a text-based household environment in which an agent receives natural-language observations and produces natural-language actions to complete domestic tasks such as cleaning, heating, cooling, and placement. Following the standard protocol used by ExpeL~\cite{zhao2024expel}, we evaluate on the 134-task subset.

The 134 tasks span six task types---\emph{Pick}, \emph{Look}, \emph{Clean}, \emph{Heat}, \emph{Cool}, and \emph{Pick2} (the underlying ALFWorld labels are \texttt{put}, \texttt{examine}, \texttt{clean}, \texttt{heat}, \texttt{cool}, \texttt{puttwo}). Table~\ref{tab:alfworld_dist} reports the per-type distribution; each fold preserves the proportion of the full set so within-fold success rates remain comparable to literature reports on the full 134-task set.

\begin{table*}[t]
\centering
\begin{tabular}{@{}lcccccccc@{}}
\toprule
 & Clean & Cool & Look & Heat & Pick & Pick2 & Total \\
\midrule
Full 134 tasks  & 31 & 21 & 18 & 23 & 24 & 17 & 134 \\
Fold 1 (Group A) & 16 & 10 & 9  & 12 & 12 & 8  & 67 \\
Fold 2 (Group B) & 15 & 11 & 9  & 11 & 12 & 9  & 67 \\
\bottomrule
\end{tabular}
\caption{ALFWorld task type distribution.}
\label{tab:alfworld_dist}
\end{table*}

\subsection{WebShop}

WebShop~\cite{yao2022webshop} is a text-based online shopping environment with a catalog of 1.18 million real Amazon products. We adopt the 100-task evaluation set released by ExpeL~\cite{zhao2024expel}, partitioned into two folds of 50 tasks each.

For each evaluated task we report two metrics. \emph{Success rate} is the fraction of episodes whose final reward equals exactly 1.0, indicating that the agent purchased a product matching all attributes specified in the instruction. \emph{Task score} is the average per-episode continuous reward, which credits partial attribute matches and can exceed binary success rate when failed purchases still match some requested attributes. Both metrics are computed using the standard WebShop reward function exactly as released by the WebShop server.

\subsubsection{Task Subtype Mapping}
\label{app:webshop_mapping}

WebShop does not provide a per-task type label, but each task carries a \texttt{key.query} field that records the Amazon catalog query used to populate its product candidates. We map each task deterministically to one of five subtypes (\emph{beauty}, \emph{clothing}, \emph{electronics}, \emph{food}, \emph{home}) by keyword-matching the query against curated category vocabularies derived from the Amazon catalog hierarchy. \emph{Beauty} covers personal-care queries such as ``bath \& bathing accessories'', ``body care'', ``eye makeup'', ``face care'', and ``dental care''. \emph{Clothing} covers apparel and footwear such as ``men's jackets'', ``women's dresses'', and ``women's mules \& clogs''. \emph{Electronics} covers consumer-electronics such as ``bluetooth headsets'', ``chargers'', and ``home theater systems''. \emph{Food} covers grocery queries such as ``alcoholic beverages'', ``cheese \& charcuterie gifts'', and ``meat \& seafood''. \emph{Home} covers furniture and home goods such as ``bedroom sets'', ``decorative pillows'', and ``kitchen islands''.

Table~\ref{tab:webshop_dist} reports the resulting distribution. The five-category decomposition supports UCE's task-type-conditioned retrieval, because memory, strategy, and workflow ECUs are bound to their generating task type.

\begin{table*}[t]
\centering
\begin{tabular}{@{}lccccccc@{}}
\toprule
 & Beauty & Clothing & Electronics & Food & Home & Total \\
\midrule
Tasks   & 31 & 14 & 21 & 15 & 19 & 100 \\
Fold 1  & 16 & 7  & 10 & 8  & 9  & 50 \\
Fold 2  & 15 & 7  & 11 & 7  & 10 & 50 \\
\bottomrule
\end{tabular}
\caption{WebShop task subtype distribution.}
\label{tab:webshop_dist}
\end{table*}

\begin{algorithm}[t]
\caption{Unified Context Evolution}
\label{alg:uce}
\begin{algorithmic}[1]
\Require Initial library $\mathcal{L}_0 = \emptyset$, agent $\pi$, eval tasks $\mathcal{E}$, collect tasks $\mathcal{Q}$, budget $B$, cycles $C$
\State $\mathbf{g}_{-1} \gets \mathbf{0}$ \Comment{per-type generation counts from prev cycle}
\For{cycle $c = 0, \dots, C{-}1$}
    \State \textbf{Phase 1 (Evaluate):} Run $\pi$ with $\mathcal{L}_c$ on $\mathcal{E}$ and record success rate $r_c$ (no fitness update)
    \State \textbf{Phase 2 (Collect):} Run $\pi$ with $\mathcal{L}_c$ on $\mathcal{Q}$, record $\mathcal{T}_c$, update ECU fitness scores, and compute collection success rate $q_c$
    \State \textbf{Phase 3 (Schedule):} $\{b^{(k)}\}_{k \in \mathcal{Z}} \gets \mathrm{KYS}(\mathcal{L}_c, q_c, B, \mathbf{g}_{c-1})$
    \Comment{Section~\ref{sec:kys}}
    \State $\mathbf{g}_c \gets \mathbf{0}$
    \For{each ECU type $k \in \mathcal{Z}$ with $b^{(k)} > 0$}
        \State \textbf{Phase 4 (Generate):} Generate $\le b^{(k)}$ candidate ECUs of type $k$ from $\mathcal{T}_c$ using the type-specific prompt with quality-feedback exemplars and reshuffle test; add survivors to $\mathcal{L}_c$ and record their count in $g_c^{(k)}$
    \EndFor
    \State \textbf{Dedup:} within each type, remove pairs with cosine similarity $> 0.85$ (keep the higher-fitness-score one)
    \State \textbf{Phase 5 (Cleanup):} drop ECUs with $u_e \geq u_{\min}$ and $f_e \leq f_{\max}$; within each (type, task\_type) group, keep only the highest-fitness-score ECU
    \State $\mathcal{L}_{c+1} \gets \mathcal{L}_c$
\EndFor
\State \Return $\mathcal{L}_C$
\end{algorithmic}
\end{algorithm}

\section{Implementation Details}
\label{app:implementation}

\subsection{Evolution Cycle Algorithm}

Algorithm~\ref{alg:uce} summarizes one complete evolution run of UCE. The algorithm preserves the read-only contract for the evaluation set and the disjoint collection set used for fitness-score updates, so the success rate $r_c$ reported per cycle is not inflated by fitness-score updates on evaluation trajectories.

\begin{table*}[t]
\centering
\setlength{\tabcolsep}{3.5pt}
\renewcommand{\arraystretch}{1.04}
\begin{tabular}{@{}>{\raggedright\arraybackslash}p{0.15\textwidth}>{\raggedright\arraybackslash}p{0.34\textwidth}>{\raggedright\arraybackslash}p{0.43\textwidth}@{}}
\toprule
\textbf{Component} & \textbf{Setting} & \textbf{Role} \\
\midrule
\textbf{Agent model} &
\makecell[l]{\texttt{gpt-4.1-mini}, $T{=}0.7$\\\texttt{max\_tokens}$=256$} &
Acting policy for each environment step. \\
\midrule
\textbf{Generator model} &
\makecell[l]{\texttt{gpt-5.2}, $T{=}0.4$\\\texttt{max\_tokens}$=1024$} &
Stable ECU extraction. \\
\midrule
\textbf{Embedding} &
\texttt{all-MiniLM-L6-v2} &
Similarity space for ECUs and tasks. \\
\midrule
\textbf{Retrieval} &
\makecell[l]{$K = 3$, $\beta = 0.15$, $\lambda = 0.1$\\dedup $= 0.85$} &
Per type top-$K$ injection, task boost, fitness-score bias, and duplicate cutoff. \\
\midrule
\textbf{KYS} &
\makecell[l]{$B = 6$, $\gamma = 0.7$\\$\alpha_{\mathrm{start}}{=}1.0$, $\alpha_{\mathrm{decay}}{=}0.1$\\$\alpha_{\min}{=}0.5$, \texttt{min\_share}$=0.10$\\$n_{\mathrm{ref}} = 3$, skill floor $= 0.1$} &
Adaptive per type generation budget. \\
\midrule
\textbf{Cleanup} &
\makecell[l]{$u_{\min} = 5$\\$f_{\max} = 0.3 / 0.2$} &
Usage gate and benchmark-specific fitness-score cutoff. \\
\midrule
\textbf{Adequacy} &
$u_{\mathrm{adq}}{=}2$, $f_{\mathrm{adq}}{=}0.5$ &
Adequacy thresholds for generation and scheduling. \\
\midrule
\textbf{Evolution loop} &
\makecell[l]{5 / 11 cycles\\30 / 20 collect tasks} &
Number of evolution cycles and collect tasks. \\
\bottomrule
\end{tabular}
\caption{Full hyperparameter list.}
\label{tab:hyperparams}
\end{table*}

\subsection{Hyperparameters}

Table~\ref{tab:hyperparams} lists all hyperparameters used in the main results.

\subsection{Control Experiments}

\begin{table}[t]
\centering\small
\setlength{\tabcolsep}{7pt}
\begin{tabular}{@{}lcc@{}}
\toprule
\textbf{Configuration} & \textbf{C0} & \textbf{Final} \\
\midrule
\multicolumn{3}{@{}l}{\textit{Generator}} \\
\texttt{gpt-5.2} (main) & 75.4 & 96.3 \\
\texttt{gpt-4.1-mini} (actor-matched) & 79.1 & 97.0 \\
\addlinespace[2pt]
\multicolumn{3}{@{}l}{\textit{Scheduling and cleanup} ($B{=}8$)} \\
KYS & 74.6 & 97.8 \\
Uniform allocation & 76.9 & 97.8 \\
Random allocation & 74.6 & 97.0 \\
Phase-5 cleanup disabled & 76.1 & 96.3 \\
\bottomrule
\end{tabular}
\caption{ALFWorld control results. Values are success rates (\%).}
\label{tab:control_results}
\end{table}

Using \texttt{gpt-4.1-mini} as both actor and generator yields a comparable final result to the main \texttt{gpt-5.2} generator. Under the matched budget, KYS and uniform allocation reach 97.8\% final success; random allocation reaches 97.0\%, and disabling Phase-5 cleanup reaches 96.3\%.

\subsection{Models and Compute}

The agent backbone (\texttt{gpt-4.1-mini}) is invoked at every environment step in both the evaluation and collection phases. The generator backbone (\texttt{gpt-5.2}) is invoked only during the Generate phase of each cycle. A full main-results run consumes on the order of tens of thousands of agent calls but only a few hundred generator calls.

All experiments run against the official OpenAI API endpoints and require no local GPU. The sentence-transformer embedding model (\texttt{all-MiniLM-L6-v2}) runs on a single CPU thread. WebShop additionally requires running its Flask server on a separate process, which loads the full 1.18-million-product Amazon catalog and a per-product search index into memory; the server needs roughly 40\,GB of RAM and several minutes of cold-start time before the agent can issue queries.

\section{Statistical Significance}
\label{app:statsig}

We test UCE's improvement over the ReAct baseline with an exact McNemar test on binary success and a 10{,}000-resample paired bootstrap on the per-task delta between Cycle~0 and the peak cycle. Table~\ref{tab:statsig} reports the results.

\begin{table}[h]
\centering\small
\setlength{\tabcolsep}{3.9pt}
\begin{tabular}{@{}lccc@{}}
\toprule
Benchmark & $\Delta$ vs.\ C0 & 95\% bootstrap CI & $p$-value \\
\midrule
ALFWorld succ.\    & $+20.9$\,pp & $[+13.4,\;+29.1]$ & $7.7{\times}10^{-7}$ \\
WebShop succ.\     & $+12.0$\,pp & $[+3.0,\;+21.0]$  & $0.017$ \\
WebShop score      & $+16.2$\,pp & $[+7.3,\;+24.7]$  & $<10^{-3}$ \\
\bottomrule
\end{tabular}
\caption{Statistical significance of UCE vs.\ the ReAct.}
\label{tab:statsig}
\end{table}

The first two rows use McNemar's exact test (binary outcome); the third uses the bootstrap (continuous reward). All three comparisons reject the null at $p < 0.05$, and all confidence intervals sit strictly above zero.

\section{Per-Cycle Evolution Trajectories}
\label{app:trajectories}

Table~\ref{tab:alfworld_cycles} reports the ALFWorld per-cycle, per-task-type success rate. Table~\ref{tab:webshop_cycles} reports the per-cycle combined success rate and task score on WebShop. Cycle~0 is the ReAct baseline, equivalent to the \textit{ReAct} row in Table~\ref{tab:main}.

\begin{table*}[t]
\centering
\setlength{\tabcolsep}{4pt}
\begin{tabular}{@{}lccccccc@{}}
\toprule
\textbf{Cycle} & Pick & Look & Clean & Heat & Cool & Pick2 & \textbf{All} \\
\midrule
C0 & 83.3 & 50.0 & 96.8 & 78.3 & 71.4 & 52.9 & 75.4 \\
C1 & 100.0 & 50.0 & 100.0 & 91.3 & 95.2 & 64.7 & 86.6 \\
C2 & 100.0 & 100.0 & 100.0 & 82.6 & 100.0 & 76.5 & 94.0 \\
C3 & 100.0 & 94.4 & 100.0 & 95.7 & 95.2 & 76.5 & 94.8 \\
\cellcolor{gray!15}\textbf{C4} & \cellcolor{gray!15}\textbf{100.0} & \cellcolor{gray!15}\textbf{100.0} & \cellcolor{gray!15}\textbf{96.8} & \cellcolor{gray!15}\textbf{95.7} & \cellcolor{gray!15}\textbf{100.0} & \cellcolor{gray!15}\textbf{82.4} & \cellcolor{gray!15}\textbf{96.3} \\
\bottomrule
\end{tabular}
\caption{ALFWorld per-cycle per-task-type success rate (\%) on the 134-task evaluation set across 5 cycles.}
\label{tab:alfworld_cycles}
\end{table*}

\begin{table*}[t]
\centering
\setlength{\tabcolsep}{4pt}
\begin{tabular}{@{}l ccccccccccc@{}}
\toprule
\textbf{Cycle} & C0 & C1 & C2 & C3 & C4 & C5 & C6 & C7 & C8 & C9 & \cellcolor{gray!15}\textbf{C10} \\
\midrule
Success rate & 30.0 & 29.0 & 36.0 & 34.0 & 35.0 & 36.0 & 36.0 & 32.0 & 38.0 & 37.0 & \cellcolor{gray!15}\textbf{42.0} \\
Task score   & 45.1 & 51.0 & 56.3 & 54.8 & 52.3 & 58.3 & 54.1 & 54.6 & 57.7 & 55.0 & \cellcolor{gray!15}\textbf{61.3} \\
\bottomrule
\end{tabular}
\caption{WebShop per-cycle success rate and task score (\%) on the 100-task evaluation set across 11 cycles.}
\label{tab:webshop_cycles}
\end{table*}

\section{Compared Methods}
\label{app:baseline_protocols}

All five baselines and three alternative actor backbones share UCE's actor model (\texttt{gpt-4.1-mini}), generator/insight model (\texttt{gpt-5.2}) where applicable, task sets, and per-episode step budgets (50 on ALFWorld, 15 on WebShop). ExpeL, ACE, AWM, ReasoningBank, and the three alternative backbones use a disjoint collection/evaluation split; cross-task knowledge is constructed only from collection trajectories, frozen for evaluation, and used for one attempt per held-out task. Reflexion has no across-task information flow and is therefore run on the merged full task set without folds, matching the convention of its original paper and of follow-up work~\cite{fu2024autoguide,kagaya2024rap,chen2024automanual,kim2025reflact}.

ExpeL~\cite{zhao2024expel} is a three-stage pipeline that gathers training-task trajectories with a reflective ReAct agent, extracts a flat pool of natural-language insights from successful trajectories and success--failure contrasts, and at evaluation time injects this pool together with $k$ nearest-neighbour successful-trajectory few-shots into the actor prompt.

Reflexion~\cite{shinn2023reflexion} gives each task up to $N$ trials, with a reflector LLM summarising each failed trial into a plan that the next trial reads. We adopt $N{=}3$, the convention shared by AutoGuide~\cite{fu2024autoguide}, ReflAct~\cite{kim2025reflact}, RAP~\cite{kagaya2024rap}, and AutoManual~\cite{chen2024automanual}; Table~\ref{tab:main} reports the resulting pass@$3$ success.

ACE~\cite{zhang2025ace} maintains an evolving playbook of structured knowledge through a generator, a reflector, and a curator. Execution feedback is converted into localized updates that add, revise, and deduplicate entries while preserving accumulated detail.

AWM~\cite{wang2024awm} abstracts successful trajectories into reusable workflows that pair a high-level goal with a generalized sequence of observations, reasoning steps, and actions. Instance-specific values are replaced with placeholders so that the workflows can guide related tasks.

ReasoningBank~\cite{ouyang2025reasoningbank} distills successful and failed trajectories into structured reasoning memories. These memories capture transferable strategies and preventive lessons that can guide later tasks.

The three actor backbones are reasoning-format variants of the same ReAct loop. NoThinking~\cite{ma2025nothinking} drops the reasoning step entirely and emits an action at every turn. Plan-and-Act~\cite{kim2025reflact}, inspired by Plan-and-Solve prompting~\cite{wang2023plansolve}, reasons only at the first turn (a high-level plan) and emits direct actions thereafter. ReflAct~\cite{kim2025reflact} prepends every turn with a one-sentence reflection on the agent's current state relative to the task goal. For each backbone, the prompt design and the procedure for deriving its few-shot examples from ReAct's base examples are taken from ReflAct~\cite{kim2025reflact}. The corresponding rows in Table~\ref{tab:main} run each backbone first with an empty UCE library and then inside the full UCE pipeline, substituting the chosen backbone for the default ReAct actor; each backbone evolves its own ECU library.

\section{Case Studies}
\label{app:case_studies}

This appendix traces four task instances across the full UCE library and the four single ECU type ablations at each configuration's peak cycle, all on the same agent backbone with the same initial observation. We cover two ALFWorld tasks---Look-Book (Case Study~\ref{case:look_book}) and Cool-Mug (Case Study~\ref{case:cool_mug})---and two WebShop tasks---Mules-Clogs (Case Study~\ref{case:mules_clogs}) and Pillow-Cover (Case Study~\ref{case:pillow_cover}). For each instance we report a summary table across all five configurations and a side-by-side trajectory comparison for the configuration pair most informative about the underlying mechanism. \ucesuccess{green} marks steps where the additional ECUs of the full library produce a successful pivot, and \ucefailure{red} marks steps where the restricted configuration enters an unrecoverable failure mode.

\begin{figure*}[t]
\centering
\refstepcounter{casestudy}\label{case:look_book}
\begin{tcolorbox}[uceframewide, title=\textbf{Case Study~\thecasestudy: Look-Book Task (ALFWorld)}]
\footnotesize

\textbf{Task instruction:} \texttt{examine the book with the desklamp.}

\centering
\begin{tabular}{lcccl}
\toprule
              & Steps & Reward & \#ECUs & Outcome \\
\midrule
Full UCE      & 22    & 1.00   & 11     & success \\
Memory only   & 50    & 0.00   & 1      & step-budget timeout \\
Strategy only & 50    & 0.00   & 1      & step-budget timeout \\
Workflow only & 50    & 0.00   & 1      & step-budget timeout \\
Skill only    & 50    & 0.00   & 1      & step-budget timeout \\
\bottomrule
\end{tabular}
\par\smallskip

\begin{minipage}[t]{0.48\linewidth}
\begin{tcolorbox}[uceinner, equal height group=examinebook]
\scriptsize\raggedright
\textbf{Workflow only}, ECUs \texttt{[wfl\_003]}\\[0.3em]
S1: \texttt{go\_to(desk 1)} (sees desklamp 1, no book).\\
\ucefailure{S2--3: \texttt{go\_to(desklamp 1)} returns ``Nothing happens'',}\\
\ucefailure{then \texttt{use(desklamp 1)} turns on the lamp with no book in hand.}\\
S4--21: search \texttt{drawer 1}--\texttt{drawer 6}, \texttt{shelf 1}--\texttt{shelf 6}, \texttt{desk 2} (no book).\\
\ucefailure{S22--36: confused loop, repeatedly tries \texttt{go\_to(book 1)} returning ``Nothing happens'' because the book has not yet been located.}\\
S37: \texttt{go\_to(bed 1)} finally finds \emph{book 1}.\\
S38: \texttt{take(book 1)}.\\
\ucefailure{S39--50: alternates \texttt{examine(book 1)} and \texttt{use(desklamp 1)} \emph{while standing at bed 1}; both fail because the agent is not co-located with the desklamp.}\\[0.3em]
Outcome: step-budget exhausted, reward 0.0.
\end{tcolorbox}
\end{minipage}\hfill
\begin{minipage}[t]{0.48\linewidth}
\begin{tcolorbox}[uceinner, equal height group=examinebook]
\scriptsize\raggedright
\textbf{Full UCE}, 11 ECUs across all four types (e.g.\ \texttt{str\_007, str\_020, wfl\_003, wfl\_006, wfl\_008, skl\_005, mem\_031})\\[0.3em]
S1--2: \texttt{go\_to(desk 2)}, \texttt{go\_to(desk 1)} (sees desklamp 1).\\
S3--8: visit \texttt{shelf 6}--\texttt{shelf 1} (no book).\\
S9--18: open and inspect \texttt{drawer 1}--\texttt{drawer 6} (no book).\\
\ucesuccess{S19: \texttt{go\_to(bed 1)} finds \emph{book 1}.}\\
\ucesuccess{S20: \texttt{take(book 1)}.}\\
\ucesuccess{S21: \texttt{go\_to(desk 1)} returns to the desklamp.}\\
\ucesuccess{S22: \texttt{use(desklamp 1)} succeeds.}\\[0.3em]
Outcome: task complete in 22 steps, reward 1.0.\\[0.3em]
After finding and taking the book, Full UCE returns to \texttt{desk 1} before using the desklamp, whereas Workflow only remains at \texttt{bed 1} and times out.
\end{tcolorbox}
\end{minipage}

\par\smallskip
\raggedright\footnotesize\textit{Analysis.} Look-Book case study on ALFWorld. All four single ECU type ablations exhaust the 50 step budget on this task. Both displayed trajectories eventually find the book on \texttt{bed 1}; Workflow only remains there and repeatedly fails to use the desklamp, while Full UCE returns to \texttt{desk 1} and solves the task in 22 steps.
\end{tcolorbox}
\end{figure*}

\begin{figure*}[t]
\centering
\refstepcounter{casestudy}\label{case:mules_clogs}
\begin{tcolorbox}[uceframewide, title=\textbf{Case Study~\thecasestudy: Mules-Clogs Task (WebShop)}]
\footnotesize

\textbf{Task instruction:} \texttt{i am looking for a grey mules \& clogs for day comfert, and price lower than 100.00 dollars.}

\centering
\begin{tabular}{lcccl}
\toprule
              & Steps & Reward & \#ECUs & Outcome \\
\midrule
Full UCE      & 7     & 1.00   & 4      & success \\
Memory only   & 15    & 0.00   & 1      & stale-search-state timeout \\
Strategy only & 15    & 0.00   & 3      & invalid-action timeout \\
Workflow only & 8     & 0.67   & 0      & partial: day-comfort mismatch \\
Skill only    & 7     & 0.67   & 1      & partial: day-comfort mismatch \\
\bottomrule
\end{tabular}
\par\smallskip

\begin{minipage}[t]{0.48\linewidth}
\begin{tcolorbox}[uceinner, equal height group=mulesclogs]
\scriptsize\raggedright
\textbf{Workflow only}, no ECU retrieved for this task\\[0.3em]
S1: \texttt{search("grey mules clogs comfort")} returns a product list.\\
S2: \texttt{think}: filter products at exactly \$100 versus those under.\\
S3: \texttt{click(B094H755LH)} opens the product page.\\
S4: \texttt{think}: pick the grey colour option.\\
S5: \texttt{click(grey)}.\\
S6: \texttt{click(7)} selects an available size required by the product page; the task specifies no size.\\
S7: \texttt{think}: ``price under \$100, attributes selected, ready to buy''.\\
\ucefailure{S8: \texttt{click(Buy Now)} returns reward 0.67.}\\[0.3em]
Outcome: mule/clog category, grey colour, and price match, but the listing misses the benchmark's day-comfort attribute; partial reward 0.67.\\[0.3em]
Workflow alone enforces the search$\rightarrow$click$\rightarrow$attribute$\rightarrow$buy template but does not verify that the candidate matches the requested day-comfort attribute.
\end{tcolorbox}
\end{minipage}\hfill
\begin{minipage}[t]{0.48\linewidth}
\begin{tcolorbox}[uceinner, equal height group=mulesclogs]
\scriptsize\raggedright
\textbf{Full UCE}, ECUs \texttt{[wfl\_001, mem\_002, str\_040, skl\_004]}\\[0.3em]
S1: \texttt{search("grey mules clogs comfort under 100")} returns a product list.\\
S2: \texttt{click(B09P33T3GG)} opens a different candidate explicitly described as comfortable and offering grey variants.\\
S3: \texttt{think}: confirm mule/clog style, day comfort, price under \$100, and variant options.\\
\ucesuccess{S4: \texttt{click(100-microsuede-grey)} picks the exact colour name from the panel rather than a generic ``grey''.}\\
S5: \texttt{click(7.5)} selects an available size required by the product page; the task specifies no size.\\
S6: \texttt{think}: day-comfort attribute and grey colour match, price \$21.98--\$23.98 under \$100.\\
\ucesuccess{S7: \texttt{click(Buy Now)} returns reward 1.0.}\\[0.3em]
Outcome: task complete in 7 steps, reward 1.0.\\[0.3em]
Unlike the Workflow-only run, the full-library run chooses a different listing that satisfies the benchmark's day-comfort component, selects grey and an available size, and completes the purchase. The reward gap therefore reflects the comfort-attribute match rather than size precision.
\end{tcolorbox}
\end{minipage}

\par\smallskip
\raggedright\footnotesize\textit{Analysis.} Mules-Clogs case study on WebShop. Memory and Strategy exhaust the budget in stale-state or invalid-action click loops, while Workflow and Skill purchase a grey, under-budget mule/clog that misses the day-comfort attribute. The full-library run retrieves all four ECU types and selects a different candidate satisfying all scored requirements, completing the purchase in seven steps.
\end{tcolorbox}
\end{figure*}

\begin{figure*}[t]
\centering
\refstepcounter{casestudy}\label{case:cool_mug}
\begin{tcolorbox}[uceframewide, title=\textbf{Case Study~\thecasestudy: Cool-Mug Task (ALFWorld)}]
\footnotesize

\textbf{Task instruction:} \texttt{cool some mug and put it in coffeemachine.}

\centering
\begin{tabular}{lcccl}
\toprule
              & Steps & Reward & \#ECUs & Outcome \\
\midrule
Full UCE      & 18    & 1.00   & 11     & success \\
Memory only   & 50    & 0.00   & 1      & step-budget timeout (stuck loop) \\
Strategy only & 19    & 1.00   & 2      & success \\
Workflow only & 19    & 1.00   & 1      & success \\
Skill only    & 50    & 0.00   & 1      & step-budget timeout \\
\bottomrule
\end{tabular}
\par\smallskip

\begin{minipage}[t]{0.48\linewidth}
\begin{tcolorbox}[uceinner, equal height group=coolmug]
\scriptsize\raggedright
\textbf{Memory only}, ECUs \texttt{[mem\_002]}\\[0.3em]
S1--9: open and inspect \texttt{cabinet 1}--\texttt{cabinet 6} (no mug found).\\
S10--12: visit \texttt{shelf 1}--\texttt{shelf 3} (no mug).\\
S13: \texttt{go\_to(coffeemachine 1)} (no mug).\\
S14--16: visit \texttt{countertop 1}--\texttt{countertop 3} (no mug).\\
S17--22: open \texttt{drawer 1}--\texttt{drawer 3} (no mug).\\
S23: \texttt{go\_to(sinkbasin 1)} (no mug).\\
\ucefailure{S24--25: \texttt{take(cup 1)} (wrong object).}\\
\ucefailure{S26--28: cool the cup, try to place it.}\\
S29--30: return the cup to \texttt{cabinet 6}.\\
\ucefailure{S31--50: 20 consecutive \texttt{go\_to(cabinet 6)}}\\
\ucefailure{$\rightarrow$ ``Nothing happens.''}\\[0.3em]
Outcome: step-budget exhausted, reward 0.0.
\end{tcolorbox}
\end{minipage}\hfill
\begin{minipage}[t]{0.48\linewidth}
\begin{tcolorbox}[uceinner, equal height group=coolmug]
\scriptsize\raggedright
\textbf{Full UCE}, 11 ECUs across all four types (e.g.\ \texttt{mem\_001, wfl\_001, str\_001, skl\_005})\\[0.3em]
S1--9: open and inspect \texttt{cabinet 1}--\texttt{cabinet 6} (no mug found).\\
S10--12: 3 consecutive \texttt{go\_to(cabinet 6)}\\
$\rightarrow$ ``Nothing happens.''\\
\ucesuccess{S13: \texttt{go\_to(countertop 1)} $\rightarrow$ finds \emph{mug 1}.}\\
\ucesuccess{S14: \texttt{take(mug 1)}.}\\
S15: \texttt{go\_to(fridge 1)}.\\
\ucesuccess{S16: \texttt{cool(mug 1, fridge 1)}.}\\
S17: \texttt{go\_to(coffeemachine 1)}.\\
\ucesuccess{S18: \texttt{put(mug 1, coffeemachine 1)}}\\
\ucesuccess{$\rightarrow$ success.}\\[0.3em]
Outcome: task complete in 18 steps, reward 1.0.
\end{tcolorbox}
\end{minipage}

\par\smallskip
\raggedright\footnotesize\textit{Analysis.} Cool mug case study on ALFWorld. All five configurations use the same task instance at their peak cycle. Full UCE finds the mug on \texttt{countertop 1} and finishes in 18 steps. Memory only and Skill only exhaust the 50 step budget, while Strategy only and Workflow only solve the task in 19 steps.
\end{tcolorbox}
\end{figure*}

\begin{figure*}[t]
\centering
\refstepcounter{casestudy}\label{case:pillow_cover}
\begin{tcolorbox}[uceframewide, title=\textbf{Case Study~\thecasestudy: Pillow-Cover Purchase Task (WebShop)}]
\small

\textbf{Task instruction:} \texttt{i want yellow machine washable batmerry summer bright decorative pillow covers, and price lower than 50.00 dollars.}

\centering
\begin{tabular}{lcccl}
\toprule
              & Steps & Reward & \#ECUs & Outcome \\
\midrule
Full UCE      & 9     & 1.00   & 3      & success \\
Memory only   & 15    & 0.00   & 3      & step-budget timeout (Invalid-action loop) \\
Strategy only & 15    & 0.00   & 3      & step-budget timeout (Invalid-action loop) \\
Workflow only & 15    & 0.00   & 1      & step-budget timeout (Invalid-action loop) \\
Skill only    & 15    & 0.00   & 1      & step-budget timeout (Invalid-action loop) \\
\bottomrule
\end{tabular}
\par\smallskip

\begin{minipage}[t]{0.48\linewidth}
\begin{tcolorbox}[uceinner, equal height group=pillow]
\scriptsize\raggedright
\textbf{Workflow only}, ECUs \texttt{[wfl\_004]}\\[0.3em]
S1: \texttt{search("yellow ... pillow covers")}\\
$\rightarrow$ product list.\\
S2: \texttt{click(B0824CBD9P)}\\
$\rightarrow$ product page.\\
S3: \texttt{think}: confirm product matches description.\\
\ucefailure{S4: \texttt{click(Features)}}\\
\ucefailure{$\rightarrow$ enters sub-tab.}\\
S5: \texttt{think}: confirm machine washable.\\
S6: \texttt{click(sunshine yellow)}.\\
S7: \texttt{click(16 x 16 inches)}.\\
\ucefailure{S8 and S9: \texttt{click(Buy Now)}}\\
\ucefailure{$\rightarrow$ ``Invalid action!'' twice}\\
\ucefailure{S10: \texttt{think}: ``Buy Now is not clickable.''}\\
\ucefailure{S11: \texttt{click(Buy Now)}}\\
\ucefailure{$\rightarrow$ ``Invalid action!''}\\
S12 to S14: three \texttt{think} actions summarizing the situation.\\
\ucefailure{S15: \texttt{click(Back to Search)}}\\
\ucefailure{step budget exhausted.}\\[0.3em]
Outcome: reward 0.0.
\end{tcolorbox}
\end{minipage}\hfill
\begin{minipage}[t]{0.48\linewidth}
\begin{tcolorbox}[uceinner, equal height group=pillow]
\scriptsize\raggedright
\textbf{Full UCE}, ECUs \texttt{[wfl\_008, mem\_057, str\_056]}\\[0.3em]
S1: \texttt{search("yellow ... pillow covers")}\\
$\rightarrow$ product list.\\
S2: \texttt{click(B0824CBD9P)}\\
$\rightarrow$ product page.\\
\ucesuccess{S3: \texttt{click(color yellow)}.}\\
\ucesuccess{S4: \texttt{click(16 x 16 inches)}.}\\
S5: \texttt{click(Features)}\\
$\rightarrow$ enters sub-tab.\\
S6: \texttt{think}: confirm machine washable.\\
S7: \texttt{click(Buy Now)}\\
$\rightarrow$ ``Invalid action!''\\
\ucesuccess{S8: \texttt{click(< Prev)}}\\
\ucesuccess{$\rightarrow$ returns to product page.}\\
\ucesuccess{S9: \texttt{click(Buy Now)}}\\
\ucesuccess{$\rightarrow$ score 1.0.}\\[0.3em]
Outcome: task complete in 9 steps, reward 1.0.
\end{tcolorbox}
\end{minipage}

\par\smallskip
\raggedright\footnotesize\textit{Analysis.} WebShop pillow cover case study. Workflow only finds the product but remains trapped inside the read only Features tab after invalid Buy Now actions. Full UCE retrieves page mechanism memory and a recovery strategy, returns with \texttt{< Prev}, and completes the purchase in nine steps.
\end{tcolorbox}
\end{figure*}

\section{Full Prompts}
\label{app:prompts}

This appendix lists every prompt UCE uses at runtime. ECU-generating prompts are color-coded by target type: Memory in blue, Strategy in red, Workflow in yellow, Skill in green; framework, role, formatting, anatomy, and feedback prompts use the default gray border.

Prompt~\ref{prompt:agent_anatomy} shows the anatomy of the agent system prompt. The system message concatenates an environment role description (Prompt~\ref{prompt:alfworld_role} for ALFWorld, Prompt~\ref{prompt:webshop_role} for WebShop), a fixed framework rules block (Prompt~\ref{prompt:framework_rules}), the four ECU sections in their canonical injection order (Prompt~\ref{prompt:ecu_headings}), and the current task description. Few-shot examples are appended afterwards as multi-turn user/assistant/tool sequences; Prompt~\ref{prompt:alfworld_fewshot} reproduces one abbreviated ALFWorld \emph{put} example. Tool actions are exposed as OpenAI function-calling schemas.

Prompt~\ref{prompt:generator_anatomy} shows the anatomy of a generator prompt. Each ECU type uses its own template, filled at Generate time with the per-type quality-feedback context (Prompt~\ref{prompt:quality_feedback}), the per-environment reshuffle test block, and the collected trajectory text. The Memory generator (Prompt~\ref{prompt:memory_gen}) extracts one reusable rule per trajectory with an outcome-conditioned guidance block (Prompt~\ref{prompt:memory_success} on success, Prompt~\ref{prompt:memory_failure} on failure). The Strategy generator (Prompt~\ref{prompt:strategy_gen}) extracts up to three decision rules per call, with an analysis-mode block that adapts to the trajectory mix (Prompt~\ref{prompt:strategy_contrastive} for mixed batches, Prompt~\ref{prompt:strategy_failure} for failure-only, Prompt~\ref{prompt:strategy_success} for success-only). The Workflow generator (Prompt~\ref{prompt:workflow_gen}) extracts a single standard step-sequence procedure from multiple successful trajectories of the same task type, and the Skill generator (Prompt~\ref{prompt:skill_gen}) extracts one or two cross-task-type sub-operation methods.

The reshuffle test distinguishes environment mechanisms from instance-specific observations and is inserted at the \texttt{\{reshuffle\_block\}} placeholder of each generator prompt. Each block contains positive (PASSES) and negative (FAILS) examples drawn from the target benchmark, with the FAILS examples tailored per ECU type. For ALFWorld the blocks are Prompts~\ref{prompt:alfworld_reshuffle_memory}--\ref{prompt:alfworld_reshuffle_skill}; for WebShop the blocks are Prompts~\ref{prompt:webshop_reshuffle_memory}--\ref{prompt:webshop_reshuffle_skill}.

\begin{figure*}[t]
\centering
\refstepcounter{prompt}\label{prompt:agent_anatomy}
\begin{tcolorbox}[uceframewide, title=\textbf{Prompt~\theprompt: Agent System Prompt anatomy}]
\small
\textbf{[Role and Domain]} You are an agent operating in a text-based interactive environment. The task type is \texttt{<task\_type>}. \\
\textbf{[Task Instructions]} Read the task description below and act step by step. Use the \texttt{think} action to plan, and use the environment-specific tool actions to interact. \\
\textbf{[Few-Shot Examples]} \emph{(category-specific, e.g.\ 1 example per WebShop subtype, 2 multi-turn examples per ALFWorld task type)} \\[6pt]

\begin{tcolorbox}[ucecomponent]
\textbf{\#\# Strategic Guidelines} \\
\emph{(top-3 retrieved \textsc{Strategy} ECUs for the current task)}
\end{tcolorbox}

\begin{tcolorbox}[ucecomponent]
\textbf{\#\# Task Workflow} \\
\emph{(top-3 retrieved \textsc{Workflow} ECUs for the current task)}
\end{tcolorbox}

\begin{tcolorbox}[ucecomponent]
\textbf{\#\# Relevant Experience} \\
\emph{(top-3 retrieved \textsc{Memory} ECUs for the current task)}
\end{tcolorbox}

\begin{tcolorbox}[ucecomponent]
\textbf{\#\# Available Skills} \\
\emph{(top-3 retrieved \textsc{Skill} ECUs, cross-task-type)}
\end{tcolorbox}

\textbf{[Task Description]} \texttt{<current task instruction>} \\
\textbf{[Initial Observation]} \texttt{<env reset output>}
\end{tcolorbox}
\end{figure*}

\begin{figure*}[t]
\refstepcounter{prompt}\label{prompt:framework_rules}
\begin{tcblisting}{uceframewide, listing only,
  title=\textbf{Prompt~\theprompt: Framework rules appended to every system prompt},
  listing options={basicstyle=\footnotesize\ttfamily, breaklines=true, breakatwhitespace=false, columns=fullflexible}}
- Call exactly ONE tool per turn.
- You may include your reasoning in the message content before calling the tool.
\end{tcblisting}
\end{figure*}

\begin{figure*}[t]
\refstepcounter{prompt}\label{prompt:alfworld_role}
\begin{tcblisting}{uceframewide, listing only,
  title=\textbf{Prompt~\theprompt: ALFWorld role + agent instructions},
  listing options={basicstyle=\footnotesize\ttfamily, breaklines=true, breakatwhitespace=false, columns=fullflexible}}
[ROLE]
You are an expert household agent.

[AGENT INSTRUCTIONS]
Interact with a household environment to complete a given task.

IMPORTANT RULES:
- You must FIRST 'go to' an object/location before you can interact with it (open, take, put, use, etc.).
\end{tcblisting}
\end{figure*}

\begin{figure*}[t]
\refstepcounter{prompt}\label{prompt:webshop_role}
\begin{tcblisting}{uceframewide, listing only,
  title=\textbf{Prompt~\theprompt: WebShop role + agent instructions},
  listing options={basicstyle=\footnotesize\ttfamily, breaklines=true, breakatwhitespace=false, columns=fullflexible}}
[ROLE]
You are an expert online shopping agent.

[AGENT INSTRUCTIONS]
Interact with an online shopping website to find and purchase a product that matches the given instruction.

IMPORTANT RULES:
- You can search for products, click on items to view details, select product options (size, color, etc.), and click 'Buy Now' to purchase.
- You must select ALL required options before clicking 'Buy Now'.
- Check that the product matches the instruction's requirements (features, price limit, etc.) before purchasing.
- You have a LIMITED step budget per task. Aim to click 'Buy Now' within the first 10 steps; do NOT spend many turns re-searching or comparing more than 3 products. If the first acceptable match meets the constraints, BUY IT --- do not over-explore looking for a marginally better option.
\end{tcblisting}
\end{figure*}

\begin{figure*}[t]
\refstepcounter{prompt}\label{prompt:ecu_headings}
\begin{tcblisting}{uceframewide, listing only,
  title=\textbf{Prompt~\theprompt: ECU section headings (injection order)},
  listing options={basicstyle=\footnotesize\ttfamily, breaklines=true}}
## Strategic Guidelines
- <Strategy ECU 1 content>
- <Strategy ECU 2 content>
- <Strategy ECU 3 content>

## Task Workflow
- <Workflow ECU 1 content>
- ...

## Relevant Experience
- <Memory ECU 1 content>
- ...

## Available Skills
- <Skill ECU 1 content>
- ...
\end{tcblisting}
\end{figure*}

\begin{figure*}[t]
\refstepcounter{prompt}\label{prompt:alfworld_fewshot}
\begin{tcblisting}{uceframewide, listing only,
  title=\textbf{Prompt~\theprompt: Abbreviated ALFWorld few-shot example (put task, first of two)},
  listing options={basicstyle=\footnotesize\ttfamily, breaklines=true, breakatwhitespace=false, columns=fullflexible}}
[user] [Example]
       You are in the middle of a room. Looking quickly around you, you see
       a cabinet 4, a cabinet 3, ..., a toilet 1, ...
       Your task is to: put some spraybottle on toilet.

[assistant content] To solve the task, I need to find and take a spraybottle,
       then put it on toilet. A spraybottle is more likely to appear in
       cabinet (1-4), countertop (1), toilet (1), sinkbasin (1-2),
       garbagecan (1). I can check one by one, starting with cabinet 1.
[assistant tool_call] go_to({"target": "cabinet 1"})
[tool] On the cabinet 1, you see a cloth 1, a soapbar 1, a soapbottle 1.

[assistant tool_call] go_to({"target": "cabinet 2"})
[tool] The cabinet 2 is closed.

[assistant tool_call] open({"target": "cabinet 2"})
[tool] You open the cabinet 2. ... you see a candle 1, and a spraybottle 2.

[assistant content] Now I find a spraybottle (2). Next, I need to take it.
[assistant tool_call] take({"object": "spraybottle 2", "source": "cabinet 2"})
[tool] You pick up the spraybottle 2 from the cabinet 2.

[assistant content] Now I take a spraybottle (2). Next, I need to put it
       in/on toilet 1.
[assistant tool_call] go_to({"target": "toilet 1"})
[tool] On the toilet 1, you see a soapbottle 2.

[assistant tool_call] put({"object": "spraybottle 2", "target": "toilet 1"})
[tool] You put the spraybottle 2 in/on the toilet 1.
\end{tcblisting}
\end{figure*}

\begin{figure*}[t]
\centering
\refstepcounter{prompt}\label{prompt:generator_anatomy}
\begin{tcolorbox}[uceframewide, title=\textbf{Prompt~\theprompt: Generator Prompt anatomy (one per ECU type)}]
\small
\textbf{[Role]} You are an experience-extraction assistant for the \texttt{<environment>} agent.

\begin{tcolorbox}[ucecomponent]
\textbf{Component 1: Type Definition and Output Format} \\
Enforces the role of the target ECU type. For example, \textsc{Memory} must output a single factual statement about an environment mechanism, \textsc{Strategy} must output IF-THEN conditional rules, \textsc{Workflow} must output a numbered step list using generalized references, \textsc{Skill} must output a multi-step sub-operation method applicable across task types.
\end{tcolorbox}

\begin{tcolorbox}[ucecomponent]
\textbf{Component 2: Quality Feedback Exemplars}\\
\textbf{POSITIVE:} highest-fitness-score ECU of this type currently in $\mathcal{L}$ \\
\textbf{NEGATIVE:} lowest-fitness-score ECU of this type currently in $\mathcal{L}$ \\
\emph{(real fitness-score values shown, guiding the generator toward higher-quality patterns through measured outcomes).}
\end{tcolorbox}

\begin{tcolorbox}[ucecomponent]
\textbf{Component 3: Reshuffle Test (mandatory)} \\
\emph{Thought experiment:} ``If all item locations were completely randomized (ALFWorld) or the product catalog were entirely replaced (WebShop), would this rule still hold?'' \\
\textbf{PASSES} (environment mechanism examples) \\
\textbf{FAILS} (instance-specific observation examples) \\
The model must argue that each candidate ECU passes the test before emitting it.
\end{tcolorbox}

\textbf{[Source Trajectories]} \texttt{<batch of Phase 2 collected trajectories with rewards>} \\
\textbf{[Output Format]} JSON list of ECU candidates, each with \texttt{when\_to\_use}, \texttt{content}, and \texttt{task\_type} fields.
\end{tcolorbox}
\end{figure*}

\begin{figure*}[t]
\refstepcounter{prompt}\label{prompt:memory_gen}
\begin{tcblisting}{ucememory, listing only,
  title=\textbf{Prompt~\theprompt: Memory generation prompt},
  listing options={basicstyle=\footnotesize\ttfamily, breaklines=true, breakatwhitespace=false, columns=fullflexible}}
You are extracting a REUSABLE RULE from a {domain_word} trajectory.

The trajectory outcome: {outcome}.

== WHAT TO EXTRACT ==
{outcome_guidance}

{reshuffle_block}

Output a single JSON object:
{
  "when_to_use": "<task type + triggering condition, e.g. 'heat tasks where the target object is on a countertop'>",
  "content": "<concrete reusable rule with action sequences, object types, or failure conditions>"
}

[Trajectory]:
{trajectory}

Output ONLY the JSON object, no additional text:
\end{tcblisting}
\end{figure*}

\begin{figure*}[t]
\refstepcounter{prompt}\label{prompt:memory_success}
\begin{tcblisting}{ucememory, listing only,
  title=\textbf{Prompt~\theprompt: Memory outcome guidance --- SUCCESS trajectory},
  listing options={basicstyle=\footnotesize\ttfamily, breaklines=true, breakatwhitespace=false, columns=fullflexible}}
This trajectory SUCCEEDED. Extract the most SURPRISING or NON-OBVIOUS finding from this experience:
- Any counter-intuitive environment behavior (an action that works without a prerequisite you'd expect, or a prerequisite you didn't expect)
- A non-obvious ordering constraint or action precondition
- An unexpected interaction between objects, containers, or appliances
If there is no surprising mechanism, extract the single most critical action or condition that determined success --- the one thing another agent is most likely to get wrong.
Do NOT output a complete step-by-step task procedure (that belongs to workflow). Focus on ONE specific insight.
Generalize from the specific objects to the task TYPE.
\end{tcblisting}
\end{figure*}

\begin{figure*}[t]
\refstepcounter{prompt}\label{prompt:memory_failure}
\begin{tcblisting}{ucememory, listing only,
  title=\textbf{Prompt~\theprompt: Memory outcome guidance --- FAILURE trajectory},
  listing options={basicstyle=\footnotesize\ttfamily, breaklines=true, breakatwhitespace=false, columns=fullflexible}}
This trajectory FAILED. Extract WHAT WENT WRONG and THE FIX:
- Identify the critical mistake or wasted steps
- Explain what the agent SHOULD have done instead
- If the agent ran out of steps, identify where it got stuck in a loop
- If the partial score is high (0.5+), the agent found a near-correct item but missed a specific attribute --- focus on WHICH attribute was wrong and HOW to verify it before purchasing
- If the partial score is 0, the agent never completed a valid purchase --- analyze whether it was stuck in navigation, search failure, or a loop
Generalize the fix so it applies to similar tasks, not just this instance. Output a SPECIFIC corrective rule, NOT a full task procedure.
\end{tcblisting}
\end{figure*}

\begin{figure*}[t]
\refstepcounter{prompt}\label{prompt:strategy_gen}
\begin{tcblisting}{ucestrategy, listing only,
  title=\textbf{Prompt~\theprompt: Strategy generation prompt},
  listing options={basicstyle=\footnotesize\ttfamily, breaklines=true, breakatwhitespace=false, columns=fullflexible}}
You are analyzing "{task_type}" task trajectories to extract reusable decision rules.

{analysis_mode}

== RULES, NOT PROCEDURES ==
Output decision rules that help at SPECIFIC CHOICE POINTS --- "do X instead of Y when Z", "skip step W because ...". Do NOT output a complete numbered step-by-step task procedure; that belongs to workflow, not strategy.

{reshuffle_block}

Output a JSON array of 1-3 rules:
[
  {
    "when_to_use": "<task type + specific triggering condition>",
    "content": "<decision rule with concrete actions or failure conditions>"
  }
]

[Trajectories]:
{trajectories}

Output ONLY the JSON array, no additional text:
\end{tcblisting}
\end{figure*}

\begin{figure*}[t]
\refstepcounter{prompt}\label{prompt:strategy_contrastive}
\begin{tcblisting}{ucestrategy, listing only,
  title=\textbf{Prompt~\theprompt: Strategy analysis mode --- contrastive (mixed success and failure)},
  listing options={basicstyle=\footnotesize\ttfamily, breaklines=true, breakatwhitespace=false, columns=fullflexible}}
You have BOTH successful and failed trajectories. This is ideal for contrastive analysis.
For each trajectory pair, find the EARLIEST decision point where the successful and failed agents diverged. Explain:
1. What the failed agent did wrong (with exact action)
2. What the successful agent did instead (with exact action)
3. The general rule that explains the difference
\end{tcblisting}
\end{figure*}

\begin{figure*}[t]
\refstepcounter{prompt}\label{prompt:strategy_failure}
\begin{tcblisting}{ucestrategy, listing only,
  title=\textbf{Prompt~\theprompt: Strategy analysis mode --- failure only},
  listing options={basicstyle=\footnotesize\ttfamily, breaklines=true, breakatwhitespace=false, columns=fullflexible}}
You have ONLY failed trajectories. Analyze the common failure patterns:
1. Identify repeated mistakes across trajectories
2. Hypothesize the correct approach based on error feedback
3. Extract rules that would prevent these failures
\end{tcblisting}
\end{figure*}

\begin{figure*}[t]
\refstepcounter{prompt}\label{prompt:strategy_success}
\begin{tcblisting}{ucestrategy, listing only,
  title=\textbf{Prompt~\theprompt: Strategy analysis mode --- success only},
  listing options={basicstyle=\footnotesize\ttfamily, breaklines=true, breakatwhitespace=false, columns=fullflexible}}
You have ONLY successful trajectories. Identify the shared NON-OBVIOUS choices and counter-intuitive actions that all successful agents performed:
1. What actions did successful agents take that a naive agent might skip or do differently?
2. Were there any steps where the environment behaved unexpectedly but all agents handled it correctly?
3. Generalize into decision rules for this task type
\end{tcblisting}
\end{figure*}

\begin{figure*}[t]
\refstepcounter{prompt}\label{prompt:workflow_gen}
\begin{tcblisting}{uceworkflow, listing only,
  title=\textbf{Prompt~\theprompt: Workflow generation prompt},
  listing options={basicstyle=\footnotesize\ttfamily, breaklines=true, breakatwhitespace=false, columns=fullflexible}}
You are extracting a STANDARD TASK PROCEDURE from multiple successful "{task_type}" task trajectories.

== WHAT TO EXTRACT ==
Identify the COMMON step sequence shared across all successful trajectories and output it as a numbered procedure template.
- Use generic references: "the target object", "the target receptacle" --- NOT specific object names or locations from any single trajectory
- Include ONLY steps that appear in ALL (or nearly all) successful trajectories --- omit exploratory detours or recovery steps unique to one trajectory
- For each step where the environment behaves counter-intuitively, add a parenthetical note explaining the mechanic (e.g. "Heat target with microwave 1 (works even when closed --- no need to open first)")

== WHAT NOT TO EXTRACT ==
- Do NOT output conditional decision rules like "if X then Y instead of Z" --- those belong to strategy, not workflow
- Do NOT specify which locations to search or in what order --- search paths vary per game instance
- Do NOT include error-recovery or backtracking steps --- output the HAPPY PATH only

{reshuffle_block}

Output a single JSON object:
{
  "when_to_use": "<task type description, e.g. 'heat tasks --- heating an object and placing it at a target location'>",
  "content": "<numbered step-by-step procedure with parenthetical mechanism notes where relevant>"
}

[Successful "{task_type}" Trajectories]:
{trajectories}

Output ONLY the JSON object, no additional text:
\end{tcblisting}
\end{figure*}

\begin{figure*}[t]
\refstepcounter{prompt}\label{prompt:skill_gen}
\begin{tcblisting}{uceskill, listing only,
  title=\textbf{Prompt~\theprompt: Skill generation prompt},
  listing options={basicstyle=\footnotesize\ttfamily, breaklines=true, breakatwhitespace=false, columns=fullflexible}}
You are identifying REUSABLE SUB-TASK METHODS that appear across multiple different {domain_word} task types.

The trajectories below come from DIFFERENT task types ({task_types}). Find operation patterns that recur across these types --- methods for sub-problems that any task might encounter.

== WHAT TO EXTRACT ==
Identify a sub-operation that appears in >= 2 different task types and describe HOW to execute it reliably.
- Focus on sub-task-level methods: searching for objects, handling containers, recovering from failed actions --- not full task procedures
- The method must be TASK-TYPE-AGNOSTIC: equally useful in {task_type_list} tasks
- Describe the method as a multi-step protocol, not as a single IF-THEN decision rule

== WHAT NOT TO EXTRACT ==
- Do NOT output a complete task procedure like "1. Find 2. Take 3. Heat 4. Put" (that belongs to workflow, not skill)
- Do NOT output a single-point decision rule like "if X then Y" (that belongs to strategy, not skill)
- Do NOT output an environment mechanism fact like "fridge doesn't need opening" (that belongs to memory, not skill)

{reshuffle_block}

Output a JSON array of 1-2 methods:
[
  {
    "when_to_use": "<functional trigger, e.g. 'when you need to locate a specific object in the environment'>",
    "content": "<multi-step method description>"
  }
]

[Trajectories from different task types]:
{trajectories}

Output ONLY the JSON array, no additional text:
\end{tcblisting}
\end{figure*}

\begin{figure*}[t]
\refstepcounter{prompt}\label{prompt:alfworld_reshuffle_memory}
\begin{tcblisting}{ucememory, listing only,
  title=\textbf{Prompt~\theprompt: ALFWorld reshuffle test --- Memory},
  listing options={basicstyle=\footnotesize\ttfamily, breaklines=true, breakatwhitespace=false, columns=fullflexible}}
== RESHUFFLE TEST (mandatory) ==
Before outputting, verify your rule passes this test:
"If ALL objects were placed in completely different locations in a new game instance, would this rule still be correct and helpful?"

PASSES (environment MECHANISM --- true across all game instances):
- "The 'cool X with fridge 1' action succeeds even when fridge is closed --- no need to 'open fridge 1' first."
- "In 'examine' tasks, the task completes when you 'use desklamp' while holding the target object --- the trigger is 'use', not 'examine'."
- "'clean X with sinkbasin 1' is a single action --- no need to put X in sinkbasin then take it back out."

FAILS (instance-specific OBSERVATION --- wrong in a different game):
- "Potato 1 is on countertop 1" --- objects are reshuffled each game.
- "Cabinet 6 cannot accept items" --- cabinet behavior varies per game.
- "Mug 2 is on cabinet 4" --- specific locations change every game.

ONLY output rules that PASS the reshuffle test.
\end{tcblisting}
\end{figure*}

\begin{figure*}[t]
\refstepcounter{prompt}\label{prompt:alfworld_reshuffle_strategy}
\begin{tcblisting}{ucestrategy, listing only,
  title=\textbf{Prompt~\theprompt: ALFWorld reshuffle test --- Strategy},
  listing options={basicstyle=\footnotesize\ttfamily, breaklines=true, breakatwhitespace=false, columns=fullflexible}}
== RESHUFFLE TEST (mandatory) ==
Every rule must pass: "If all objects were in different locations in a new game, would this rule still be correct and helpful?"

PASSES (environment mechanism --- decision rules):
- "'cool X with fridge 1' works even when fridge is closed --- do NOT waste steps opening the fridge before cooling."
- "'heat X with microwave 1' works even when microwave is closed --- skip the 'open microwave' step."
- "In 'examine' tasks, the completion trigger is 'use desklamp', NOT 'examine [object]'. Attempting to examine the object directly will not complete the task."

FAILS (DO NOT output):
- "The egg is in the fridge" / "Cabinet 3 contains a mug" (instance-specific)
- "Explore efficiently" / "plan ahead" (generic, any LLM knows this)
- "1. Find object 2. Take it 3. Heat it 4. Put it" (step-by-step procedure --- belongs to workflow, not strategy)
- "Search method: visit each receptacle, open closed ones, check contents, move to next" (sub-task method --- belongs to skill, not strategy)
\end{tcblisting}
\end{figure*}

\begin{figure*}[t]
\refstepcounter{prompt}\label{prompt:alfworld_reshuffle_workflow}
\begin{tcblisting}{uceworkflow, listing only,
  title=\textbf{Prompt~\theprompt: ALFWorld reshuffle test --- Workflow},
  listing options={basicstyle=\footnotesize\ttfamily, breaklines=true, breakatwhitespace=false, columns=fullflexible}}
== RESHUFFLE TEST (mandatory) ==
Verify: "If all objects were renamed and placed in completely different locations, would this step sequence still be the correct procedure for {task_type} tasks?"

PASSES (generalised procedure with mechanism notes):
- "Heat: 1. Find the target object (search receptacles until found) 2. Take it 3. Go to microwave 1 4. Heat target with microwave 1 (works even when closed) 5. Go to target receptacle 6. Put target in/on target receptacle"
- "Examine: 1. Find the target object 2. Take it 3. Go to desklamp 1 4. Use desklamp 1 (task completes on 'use' while holding the object --- 'examine' alone does NOT complete it)"

FAILS (DO NOT output):
- "1. Go to countertop 1 2. Take potato 1 3. Go to microwave 1 ..." --- uses specific object names and locations from one game instance
- "1. Find the object 2. Do the required action 3. Place it" --- too vague to be actionable
\end{tcblisting}
\end{figure*}

\begin{figure*}[t]
\refstepcounter{prompt}\label{prompt:alfworld_reshuffle_skill}
\begin{tcblisting}{uceskill, listing only,
  title=\textbf{Prompt~\theprompt: ALFWorld reshuffle test --- Skill},
  listing options={basicstyle=\footnotesize\ttfamily, breaklines=true, breakatwhitespace=false, columns=fullflexible}}
== RESHUFFLE TEST (mandatory) ==
Verify: "Would this method work identically in a completely different game instance with different object placements?"

PASSES (cross-type reusable method):
- "Systematic object search: Visit each receptacle in the room. For closed containers (cabinet/drawer), open before checking contents. When the target appears in the observation, immediately take it. If 'take' returns 'Nothing happens', move to the next receptacle."
- "Failure recovery for 'Nothing happens': (1) Verify co-location by doing 'go to <target>'. (2) If the target is a container, open it. (3) Retry the action once. (4) If still failing, switch to a different instance or location."

FAILS (DO NOT output):
- "Heat task: 1. Find object 2. Take 3. Microwave 4. Put" (task-level procedure --- belongs to workflow)
- "If 'put' returns Nothing happens, open the container" (single IF-THEN rule --- belongs to strategy)
- "The fridge doesn't need to be opened for cooling" (environment fact --- belongs to memory)
\end{tcblisting}
\end{figure*}

\begin{figure*}[t]
\refstepcounter{prompt}\label{prompt:webshop_reshuffle_memory}
\begin{tcblisting}{ucememory, listing only,
  title=\textbf{Prompt~\theprompt: WebShop reshuffle test --- Memory},
  listing options={basicstyle=\footnotesize\ttfamily, breaklines=true, breakatwhitespace=false, columns=fullflexible}}
== RESHUFFLE TEST (mandatory) ==
Before outputting, verify your rule passes this test:
"If the entire product catalog changed (different products, prices, and options), would this rule still be correct and helpful?"

PASSES (environment MECHANISM --- true across all shopping sessions):
- "You must select ALL required options (size, color, etc.) before clicking 'Buy Now', or the purchase will receive a score of 0."
- "The 'think[...]' action does not change the page state --- use it freely to plan without consuming navigation steps."
- "Clicking 'Back to Search' from any product page returns to the initial search page, allowing you to start a completely new query."

FAILS (instance-specific OBSERVATION --- wrong with different products):
- "B078GWRC1J is the best match for deodorant" --- product IDs and inventory change across sessions.
- "The first search result is always the cheapest" --- result ordering varies by query and session.
- "Earth Mama deodorant costs $10.99" --- prices are session-specific.

ONLY output rules that PASS the reshuffle test.

== BUDGET CONSTRAINT ==
Each episode has a strict per-episode step budget. Memory ECUs that add verification clicks consume budget; memory ECUs that warn against unsalvageable states save budget.

PREFER rules of the form "Treat <state> as <non-purchasable / stale>; do NOT do <action> from this state":
- "Treat Attributes/Description/Features sub-tabs as read-only views; 'Buy Now' from these tabs returns Invalid Action. Use '< Prev' to return to the main product page before purchasing."
- "Treat the generic [Search]-only screen as a reset state; do NOT click product IDs or variants here. Click 'Search' to reload the result list."

AVOID rules that require additional verification clicks before Buy Now:
- "Before purchasing, verify every constraint by opening Description, Features, Attributes, Reviews." Each sub-tab click is a step; this adds 4 steps before the actual purchase.
- "Always read at least three Reviews before deciding." Each review click is a step.

If verification is necessary, encode it as a think[...] check rather than as additional click actions.
\end{tcblisting}
\end{figure*}

\begin{figure*}[t]
\refstepcounter{prompt}\label{prompt:webshop_reshuffle_strategy}
\begin{tcblisting}{ucestrategy, listing only,
  title=\textbf{Prompt~\theprompt: WebShop reshuffle test --- Strategy},
  listing options={basicstyle=\footnotesize\ttfamily, breaklines=true, breakatwhitespace=false, columns=fullflexible}}
== RESHUFFLE TEST (mandatory) ==
Every rule must pass: "If the product catalog changed completely, would this rule still be correct and helpful?"

PASSES (environment mechanism --- decision rules):
- "If the initial search returns irrelevant results, click 'Back to Search' and try a shorter, more specific query rather than browsing multiple pages of poor results."
- "Always verify the item's price against the budget constraint in the instruction BEFORE clicking 'Buy Now' --- buying an over-budget item scores 0."
- "When a product has required options (color, size, etc.), select ALL of them before clicking 'Buy Now'. Missing any option results in a failed purchase."

FAILS (DO NOT output):
- "Search for 'bright citrus deodorant' to find the target product" (instance-specific query)
- "Always pick the cheapest option" (oversimplified, ignores feature matching)
- "1. Search 2. Click product 3. Select options 4. Buy" (step-by-step procedure --- belongs to workflow, not strategy)
- "Check every product on every search page systematically" (sub-task method --- belongs to skill, not strategy)

== BUDGET CONSTRAINT ==
Each episode has a strict per-episode step budget. Strategy ECUs that delay commitment by adding verification or comparison clicks consume budget; strategies that detect a bad state and immediately abandon save budget.

PREFER strategies of the form "IF <bad signal> THEN <abandon / reset / move on>":
- "IF clicking 'Buy Now' returns 'Invalid action!' AND you have already retried once from this listing, THEN abandon this listing and open a different search result."
- "IF a click on a result ASIN does not navigate to a product page, THEN do not retry the same ASIN; click 'Search' to reload the result list."

AVOID strategies that require additional verification or comparison clicks before Buy Now:
- "Always click 'Description' and 'Features' to confirm the product before Buy Now." This consumes 2 sub-tab clicks per attempt.
- "Compare at least three candidate products by opening each one before deciding." Multi-product comparison consumes 6+ steps.

If a strategy must guard against a constraint, encode the check as a think[...] verification rather than as additional click actions.
\end{tcblisting}
\end{figure*}

\begin{figure*}[t]
\refstepcounter{prompt}\label{prompt:webshop_reshuffle_workflow}
\begin{tcblisting}{uceworkflow, listing only,
  title=\textbf{Prompt~\theprompt: WebShop reshuffle test --- Workflow},
  listing options={basicstyle=\footnotesize\ttfamily, breaklines=true, breakatwhitespace=false, columns=fullflexible}}
== RESHUFFLE TEST (mandatory) ==
Verify: "If the product catalog changed completely, would this step sequence still be the correct procedure for shopping tasks?"

PASSES (generalised procedure with mechanism notes):
- "Shopping: 1. Search with keywords from the instruction 2. Identify a matching product from results (check title + price) 3. Click the product ASIN to view details 4. Select all required options (color, size, etc.) 5. Verify price is within budget 6. Click 'Buy Now' to complete purchase"

FAILS (DO NOT output):
- "1. Search 'deodorant' 2. Click B078GWRC1J 3. Select 'bright citrus' 4. Click 'Buy Now'" --- uses specific product IDs and options from one session
- "1. Search 2. Buy the first result" --- too vague to be actionable
\end{tcblisting}
\end{figure*}

\begin{figure*}[t]
\refstepcounter{prompt}\label{prompt:webshop_reshuffle_skill}
\begin{tcblisting}{uceskill, listing only,
  title=\textbf{Prompt~\theprompt: WebShop reshuffle test --- Skill},
  listing options={basicstyle=\footnotesize\ttfamily, breaklines=true, breakatwhitespace=false, columns=fullflexible}}
== RESHUFFLE TEST (mandatory) ==
Verify: "Would this method work identically with a completely different product catalog?"

PASSES (cross-session reusable method):
- "Efficient product evaluation: Read the title and price of each search result. Skip products that clearly don't match (wrong category, over budget, missing key features). Only click on products whose title partially matches the instruction to check their detailed options and price."
- "Search query refinement: If the first search returns irrelevant results, go back and try (1) fewer keywords focusing on the core product type, (2) different synonyms, or (3) brand name if mentioned in the instruction."

FAILS (DO NOT output):
- "Shopping procedure: Search -> Click -> Select -> Buy" (task-level procedure --- belongs to workflow)
- "If price exceeds budget, go back and find cheaper option" (single IF-THEN rule --- belongs to strategy)
- "'Buy Now' completes the task and reveals the score" (environment fact --- belongs to memory)

== BUDGET CONSTRAINT ==
Each episode has a strict per-episode step budget. Skill ECUs that ask the agent to "scan all variants" or "open every sub-tab to verify" consume budget; recovery protocols whose terminal step is "abandon and try next listing" save budget.

PREFER recovery / navigation skills whose terminal step is "abandon and move to the next candidate" or "reset to a known-good view in ONE step":
- "Invalid-action recovery: 1) identify which view you are on; 2) return to the main product page in ONE step (use '< Prev' or re-click the product from search results); 3) re-try the action ONCE from the restored view; 4) if it still fails, abandon this listing and open a different search result."
- "Stale-state escape: when the page becomes a generic [Search] control with no result list, click 'Search' once to reload; if the result list does not return after one reload, issue a fresh search query rather than continuing to click."

AVOID skill protocols that require an analysis pass before any action:
- "Scan all variant selectors; list which are constrained vs free; locate each required value; click them in a stable order before 'Buy Now'." This consumes 4-6 steps before the first commitment.
- "Compare top-3 search results by opening each, recording prices and option ranges, then return to the best candidate." Multi-product comparison protocols inflate the step count beyond the per-episode budget on most environments.

If a skill must verify constraints, encode the verification as a think[...] step inside the protocol rather than as additional click actions.
\end{tcblisting}
\end{figure*}

The quality-feedback context block (Prompt~\ref{prompt:quality_feedback}) is prepended to each generator prompt and contains up to three components: a quality reference drawn from the highest- and lowest-fitness-score ECUs of the same type, a library-coverage overview, and a coverage-gap nudge for under-covered task types. The reference component is omitted in the first few cycles when the library is still empty.

\begin{figure*}[t]
\refstepcounter{prompt}\label{prompt:quality_feedback}
\begin{tcblisting}{uceframewide, listing only,
  title=\textbf{Prompt~\theprompt: Quality-feedback context template (one block per generator call)},
  listing options={basicstyle=\footnotesize\ttfamily, breaklines=true, breakatwhitespace=false, columns=fullflexible}}
== QUALITY REFERENCE (from previously generated ECUs) ==
HIGH quality (fitness={best_fitness}, used {best_usage_count} times):
  when_to_use: "{best_when_to_use}"
  content: "{best_content}"
  Why good: Captures a specific, counter-intuitive mechanism that the agent cannot learn from few-shot examples alone.

LOW quality (fitness={worst_fitness}, used {worst_usage_count} times):
  when_to_use: "{worst_when_to_use}"
  content: "{worst_content}"
  Why bad: Too generic or duplicates information already available in the few-shot demonstrations.

Generate an ECU that is more like the HIGH quality example.

== CURRENT LIBRARY STATUS ==
  memory: {n_memory} ECUs ({memory_task_types})
  strategy: {n_strategy} ECUs ({strategy_task_types})
  workflow: {n_workflow} ECUs ({workflow_task_types})
  skill: {n_skill} ECUs ({skill_task_types})

Gap for {ecu_type}: task types {uncovered_task_types} have no coverage yet. Prioritize generating knowledge for uncovered types if the trajectory data supports it.
\end{tcblisting}
\end{figure*}

\section{Statement on AI Assistance}
\label{app:ai_assistants}

We used general-purpose AI assistants for coding support (completion, refactoring, debugging) and for sentence-level writing polish. All algorithmic design, experimental decisions, results, and analyses are the authors' own.

\end{document}